\documentclass[final]{cvpr}

\usepackage{times}
\usepackage{epsfig}
\usepackage{graphicx}
\usepackage{amsmath}
\usepackage{amssymb}
\usepackage{cite}
\usepackage{soul}
\usepackage{multirow}
\usepackage{gensymb}
\usepackage{bm}
\usepackage{makecell}
\usepackage{capt-of}

\newcommand\blfootnote[1]{%
	\begingroup
	\renewcommand\thefootnote{}\footnote{#1}%
	\addtocounter{footnote}{-1}%
	\endgroup
}

\newcommand{\subtitle}[1]{{\noindent}{\textbf{#1}}.}

\newcommand{\ping}[1]{\textcolor[rgb]{0.8,0.2,0.2}{[Ping: #1]}}

\newcommand{\zp}[1]{\textcolor[rgb]{0.0,0.0,1.0}{[Zhaopeng: #1]}}
\newcommand{\ziqian}[1]{#1}

\usepackage[pagebackref=true,breaklinks=true,colorlinks,bookmarks=false]{hyperref}

\def\cvprPaperID{4971} 
\def\confYear{CVPR 2021}
\begin{document}

\title{Riggable 3D Face Reconstruction via In-Network Optimization}

\author{Ziqian Bai$^{1}$ \ \ Zhaopeng Cui$^{2*}$ \ \ Xiaoming Liu$^{3}$ \ \ Ping Tan$^{1*}$ \\
$^{1}$ Simon Fraser University \ \ $^{2}$ State Key Lab of CAD \& CG, Zhejiang University \\ $^{3}$ Michigan State University \\
{\tt\small \{ziqianb, pingtan\}@sfu.ca,  zhpcui@zju.edu.cn, liuxm@cse.msu.edu
}
}

\twocolumn[{%
\renewcommand\twocolumn[1][]{#1}%
\maketitle
\begin{center}
    \vspace{-1.0em}
    \centering
    \includegraphics[width=\linewidth]{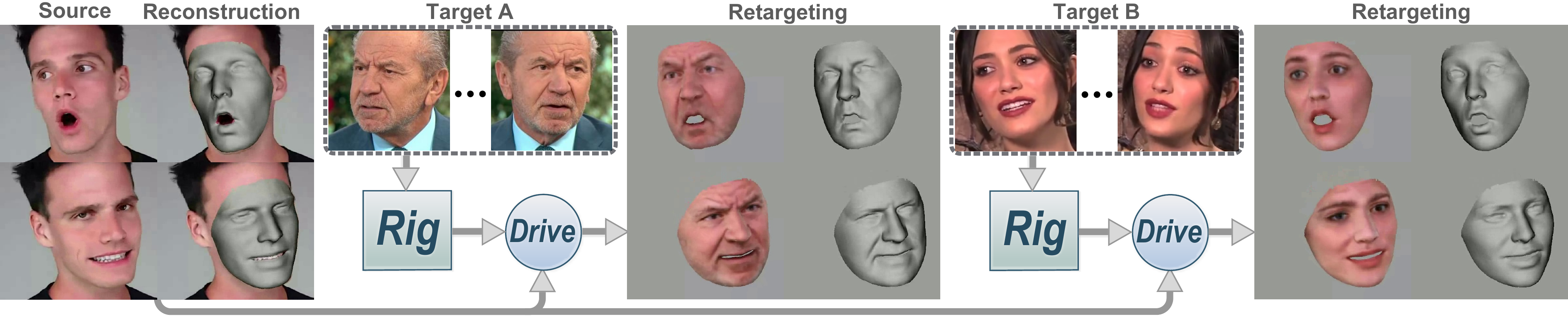}
    \captionof{figure}{Our method estimates personalized face rigs and per-image reconstructions from monocular images with good reconstruction quality and supports video retargeting to different actors.
    The code is available at \href{https://github.com/zqbai-jeremy/INORig}{\color{cyan}{https://github.com/zqbai-jeremy/INORig}}.}
    \label{fig:teaser}
    
\end{center}%
}]
\thispagestyle{empty}

\begin{abstract}
   This paper presents a method for riggable 3D face reconstruction from monocular images, which jointly estimates a personalized face rig and per-image parameters including expressions, poses, and illuminations. To achieve this goal, we design \ziqian{an end-to-end trainable network embedded with a differentiable in-network optimization. The network}
first parameterizes the face rig as a compact latent code with a \ziqian{neural}
decoder, and then estimates the latent code as well as per-image parameters via a learnable optimization. By estimating a personalized face rig, our method goes beyond static reconstructions and enables downstream applications such as video retargeting. 
\ziqian{In-network} optimization 
\ziqian{explicitly} enforces constraints derived from the first principles, thus introduces additional priors than regression-based methods. Finally, data-driven priors from deep learning are utilized to constrain the ill-posed monocular setting and ease the optimization difficulty. Experiments demonstrate that our method achieves SOTA reconstruction accuracy, reasonable robustness and generalization ability, and supports standard face rig applications.\blfootnote{$^{*}$Corresponding authors}

\end{abstract}

\setlength{\abovedisplayskip}{3pt}
\setlength{\belowdisplayskip}{3pt}
\section{Introduction}
3D face reconstruction has been an important research topic due to the increasing demands on 3D face understanding in fields like AR/VR, communication, games, and security. 
Some approaches go beyond merely estimating static reconstructions and aim to reconstruct face rigs, which are personalized parametric models that can produce 3D faces under different expressions of a specific person. The rig can either be used on character animations such as face retargeting and voice puppetry, or on 3D face tracking serving as a personalized prior to ease the tracking difficulty.

When 3D data is available, various approaches \cite{li2010example,bouaziz2013online,li2013realtime} have been proposed to automatically reconstruct face rigs in the forms of blendshapes. Progress has also been made to develop more sophisticated rigs based on anatomical constraints \cite{wu2016anatomically} and deep neural networks \cite{lombardi2018deep,wu2018deep} to faithfully capture facial details. However, these methods heavily depend on the 3D data provided by specialized equipment such as dense camera/lighting arrays and depth sensors, which limits the application realms.

To release the restricted hardware requirements, methods were enhanced to work on monocular imagery. Given the ill-posedness of monocular reconstruction, algorithms usually use a low dimensional parametric face model as priors, \eg, 3D morphable model (3DMM) \cite{blanz1999morphable} and multi-linear model \cite{vlasic2006face,cao2013facewarehouse}, whose parameters are estimated via the analysis-by-synthesis optimization~\cite{zollhofer2018state,egger20203d,hu2017avatar,garrido2016reconstruction}. Additional components such as corrective basis~\cite{garrido2016reconstruction}, shading-based dynamic details~\cite{ichim2015dynamic,garrido2016reconstruction}, image-based representation~\cite{cao2016real}, as well as hair~\cite{cao2016real,hu2017avatar} and other secondary components \cite{ichim2015dynamic} are adopted to further personalize the estimated rig. However, these approaches may assume specific properties of the input, \eg, requiring the subject to be static and in the neutral pose for a portion of the input~\cite{ichim2015dynamic,cao2016real}; need manual intervention~\cite{ichim2015dynamic}; and are often inefficient~\cite{garrido2016reconstruction}. 


The recent boom in deep learning also advanced monocular 3D face reconstruction. Various learning-based methods were proposed to regress face model parameters or face shapes \cite{tuan2017regressing,richardson20163d,sela2017unrestricted}, learn with novel supervisions \cite{deng2019accurate,genova2018unsupervised,sanyal2019learning}, build better face models \cite{on-learning-3d-face-morphable-model-from-in-the-wild-images,towards-high-fidelity-nonlinear-3d-face-morphable-model,tewari2018self,tewari2019fml}, as well as integrate with traditional multi-view geometry~\cite{bai2020deep}. Nevertheless, these methods mainly focus on static reconstructions and fail to produce personalized face rigs. Very recently, Chaudhuri \etal~\cite{chaudhuri2020personalized} used neural networks for regressing blendshape face rigs from monocular images. Despite the appealing textures produced by their method, the estimated 3D geometry, which is an important aspect for 3D reconstruction, still has considerable room for improvement.


In this paper, we propose a \ziqian{monocular} riggable 3D face reconstruction algorithm.
The riggable reconstruction consists of a personalized face rig \ziqian{and per-image parameters including expressions, poses, and illuminations.}
\ziqian{Our method is an end-to-end trainable network embedded with a differentiable in-network optimization. Two modules are involved. One is a neural decoder conditioned on the input images to}
parameterize the face rig into a latent code (termed as rig code) to control the person-specific aspects (\eg identity).
\ziqian{The other is a learnable optimization that estimates}
the rig code \ziqian{and}
the per-image parameters.

\ziqian{Our main novelty is the {\it integration} of deep learning and optimization for face rig. In contrast to prior static reconstruction methods \cite{bai2020deep}, our riggable}
reconstruction 
can be re-animated by another face or even voices, enabling extra applications such as face retargeting and voice puppetry.
\ziqian{Different from previous learning-based methods \cite{chaudhuri2020personalized} that {\it directly regress} rig parameters, our in-network optimization iteratively solves rig parameters with {\it explicit constraints} governed by the first-principles}
(\eg multi-view consistency, landmark alignment, and photo-metric reconstruction),
achieving better geometry accuracy and \ziqian{good} data generalization.
\ziqian{Unlike traditional optimizations \cite{ichim2015dynamic,garrido2016reconstruction} using hand-crafted priors, we adopt a learned deep rig model and a learned optimization to leverage deep priors}
to constrain the ill-posedness and ease the hardness of the optimization.
Our method is able to achieve state-of-the-art (SOTA) reconstruction accuracy, reasonable robustness and generalization ability, and can be used in 
 standard face rig applications as demonstrated in experiments.

\section{Related Works}
\subtitle{Personalized Modeling with 3D Inputs}
Traditionally, blendshapes are typical choices to represent a personalized face model/rig, which are 
expressive $3$D shapes that can be linearly combined to \ziqian{get}
novel expressions \cite{lewis2014practice}. Due to the tremendous effort of manually creating blendshapes, approaches \cite{li2010example,bouaziz2013online,li2013realtime,hsieh2015unconstrained} have been proposed to automate the process by adaptively updating \ziqian{the initial}
blendshapes or adding correctives based on 
3D inputs such as example meshes or \ziqian{depths.}
Progress has also been made to develop more sophisticated models. Wu \etal \cite{wu2016anatomically} proposed a local face model with anatomical constraints. 
Deep neural networks \cite{lombardi2018deep,wu2018deep} are also employed to capture person-specific geometry and appearance details.
However, the application scenarios of these methods are largely limited by the dependency on 3D data, which requires specialized equipments such as dense camera/lighting arrays and depth sensors.

Our method only needs monocular RGB images as inputs, thus eliminates the dependency of bulky equipments.

\subtitle{Traditional Optimization}
Methods were proposed to reconstruct personalized face rigs from monocular RGB \ziqian{data}.
Usually, some parametric face models, such as 3D morphable model (3DMM) \cite{blanz1999morphable,blanz2003face,adaptive-3d-face-reconstruction-from-unconstrained-photo-collections} or multi-linear blendshapes \cite{vlasic2006face,cao2013facewarehouse}, are used as priors to constrain the ill-posed \ziqian{problem},
while the model parameters are computed by various types of optimization~\cite{zollhofer2018state}. 
Different algorithms were designed by extending this basic fitting pipeline. Ichim \etal \cite{ichim2015dynamic} fit a $3$DMM to a structure-from-motion reconstruction and personalize the expressions by updated blendshapes and a learned detail map regressor.
Garrido \etal \cite{garrido2016reconstruction} proposed a multi-layer representation to reconstruct personalized face rigs from monocular RGB videos in a fully automatic fashion. 
\ziqian{People} \cite{cao2016real,hu2017avatar} \ziqian{also added}
hair and other secondary components 
for more realistic face rigs. However, the \ziqian{$3$DMM or multi-linear models}
involved usually cannot capture accurate face geometry due to their limited \ziqian{capacity.}
Though this can be alleviated by
further adaptations or correctives, these methods usually contain a long pipeline with
heavily engineered components, which may require manual intervention \cite{ichim2015dynamic} and are often fragile and \ziqian{inefficient.}

Recently, deep learning becomes a potential solution to address these issues. Our method leverages the powerful and more general data-driven priors captured by networks to constrain the ill-posedness and ease the  optimization.  

\begin{figure*}[h!]
	\centering
    \includegraphics[width=0.85\linewidth]{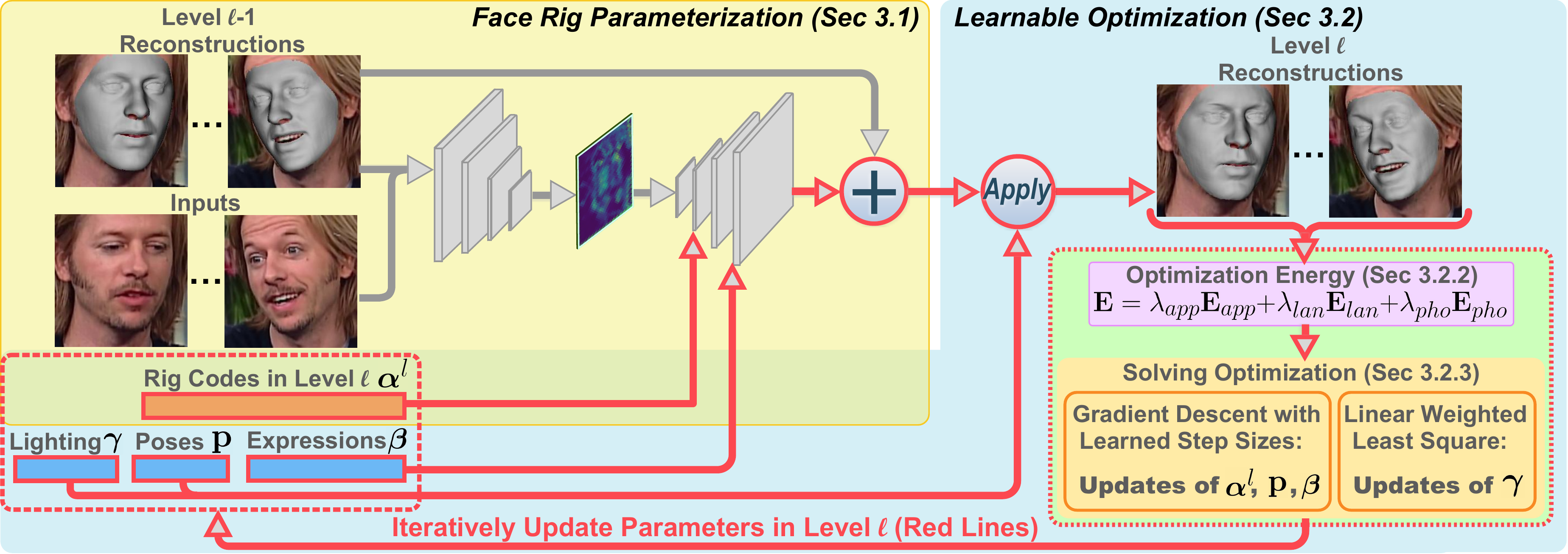}
    \caption{The single level illustration of our method, which is repeated $3$ times to form a multi-level scheme. Two modules are involved: (1) \textbf{\textit{Face Rig Parameterization}} that parameterizes the face rig into an optimisable latent code $\bm{\alpha}^l$ to control the person-specific aspects (\eg identity) via a neural decoder; (2) An end-to-end \textbf{\textit{Learnable Optimization}} to iteratively update the rig code $\bm{\alpha}^l$ and the per-image 
    parameters including expressions, poses, and illuminations.}
	\label{fig:overview}
	\vspace{-1.25em}
\end{figure*}

\subtitle{Learning-based Methods}
Plenty of deep learning methods were designed to regress $3$D shapes or face model parameters \cite{tuan2017regressing,richardson20163d,sela2017unrestricted}, learn with \ziqian{only $2$D images~\cite{deng2019accurate,genova2018unsupervised} and identity~\cite{sanyal2019learning} supervisions,}
learn better face models from in-the-wild data \cite{on-learning-3d-face-morphable-model-from-in-the-wild-images,towards-high-fidelity-nonlinear-3d-face-morphable-model,tewari2018self,tewari2019fml}, as well as integrate with traditional multi-view geometry~\cite{bai2020deep}. However, most of them focus on static reconstructions instead of personalized face rigs. Very recently, Yang \etal \cite{yang2020facescape} proposed to regress riggable displacement maps acting as textures of a bilinear blendshape model fitted by traditional optimizations. \ziqian{Though the displacement maps give better visual quality, they cannot address the limited capacity}
of linear models in terms of geometry accuracy. 
Chaudhuri \etal \cite{chaudhuri2020personalized} proposed to use networks for regressing blendshape face rigs from monocular images in a self-supervised manner. 
Despite the appealing textures produced by their algorithm, their estimated 3D geometry, which is an important aspect for 3D reconstruction, still has considerable room for improvement.

Instead of direct regression, our method 
\ziqian{uses in-network} optimization \ziqian{governed by the first-principles.}
This \ziqian{extra constraint, together with the learned deep priors,}
offer the potential to improve geometry accuracy and generalization, while address the limited  capacity of linear face models.


\section{Method}
Given $N$ monocular RGB images $\{\mathbf{I}_i\}_{i=1}^{N}$ of a person (\ie unsynchronized images taken under different views and expressions), our method estimates riggable 3D face reconstructions composed of a personalized face rig $Rig(\cdot)$ as well as per-image parameters $\{\mathbf{x}_i = (\bm{\beta}_i, \mathbf{p}_i, \bm{\gamma}_i)\}_{i=1}^{N}$ including expressions $\bm{\beta}_i$, poses $\mathbf{p}_i$, and illuminations $\bm{\gamma}_i$. The per-image 3D reconstruction can be obtained by combining the estimated face rig and per-image parameters.

Our framework adopts a $3$-level scheme to perform the reconstruction in a coarse-to-fine manner. For each level $l$, there are mainly $2$ modules (see \figref{fig:overview}): (1) Face Rig Parameterization (\secref{sec:rig_decoder}): An image-conditioned network decoder to parameterize the face rig updates at level $l$ into an optimizable latent code $\bm{\alpha}^l$; (2) Learnable Optimization (\secref{sec:opt}): An end-to-end learnable optimization to iteratively update the rig code $\bm{\alpha}^l$ and per-image parameters $\{\mathbf{x}_i\}_{i=1}^{N}$. 
Finally, our model is trained with registered ground truth 3D scans in a supervised manner (\secref{sec:loss}). 


\subsection{Face Rig Parameterization}
\label{sec:rig_decoder}
The face rig is a parametric model that takes in an expression parameter $\bm{\beta}$ and outputs a colored 3D face mesh corresponding to the input expression, denoted as $(\mathbf{V}, \mathbf{A}) = Rig(\bm{\beta})$ where $\mathbf{V}$ is the mesh vertices and $\mathbf{A}$ is the albedo colors. To model the face rig $Rig(\cdot)$, multiple approaches have been proposed such as using a set of blendshapes \cite{chaudhuri2020personalized}, a neural network \cite{wu2018deep}, or multi-layer representations \cite{garrido2016reconstruction}. However, these models are usually hard to 
be optimized \footnote{Here ``optimize" refers to optimizing the model parameters over current input images, but not \ziqian{over the whole dataset (\ie training).}
} due to the ill-posedness of monocular 3D reconstruction. Inspired by previous works \cite{bloesch2018codeslam,tang2019ba,bai2020deep}, we propose to learn a compact and optimisable latent code $\bm{\alpha}$ via the neural network to parameterize the face rig for constraining the ill-posed problem via data-driven priors.

Specifically, we design a neural decoder that takes in the latent code $\bm{\alpha}$ (termed as ``rig code") and the expression parameter $\bm{\beta}$, and outputs the colored mesh $(\mathbf{V}, \mathbf{A})$. We make the decoder conditioned on the input images $\{\mathbf{I}_i\}$ as in \cite{bloesch2018codeslam,tang2019ba,bai2020deep} to better leverage the visual clues. Note that we also need some sort of initial/intermediate reconstructions $\{\mathbf{\hat{V}}^{old}_i\}$ (\ie per-image meshes of level $l-1$ shown in \figref{fig:overview}) to map the image information into UV space. We will describe how to get the initial/intermediate reconstructions $\{\mathbf{\hat{V}}^{old}_i\}$ in \secref{sec:solve_opt}. Formally, we have
\begin{align}
    (\mathbf{V}, \mathbf{A}) = Rig(\bm{\beta} ; \bm{\alpha}, \{\mathbf{I}_i\}, \{\mathbf{\hat{V}}^{old}_i\}).
\label{eq:rig_dec}
\end{align}
Inside the rig, there are mainly $4$ components (each has $3$ levels): image feature extraction, neutral shape decoding, expression deformation decoding, and albedo decoding, which will be described one by one. For each component (except Albedo), we will firstly describe its single level version, then  its generalization to the $3$-level scheme. More details can be found in the supplementary material.

\begin{figure*}[!t]
	\centering
    \includegraphics[width=0.9\linewidth]{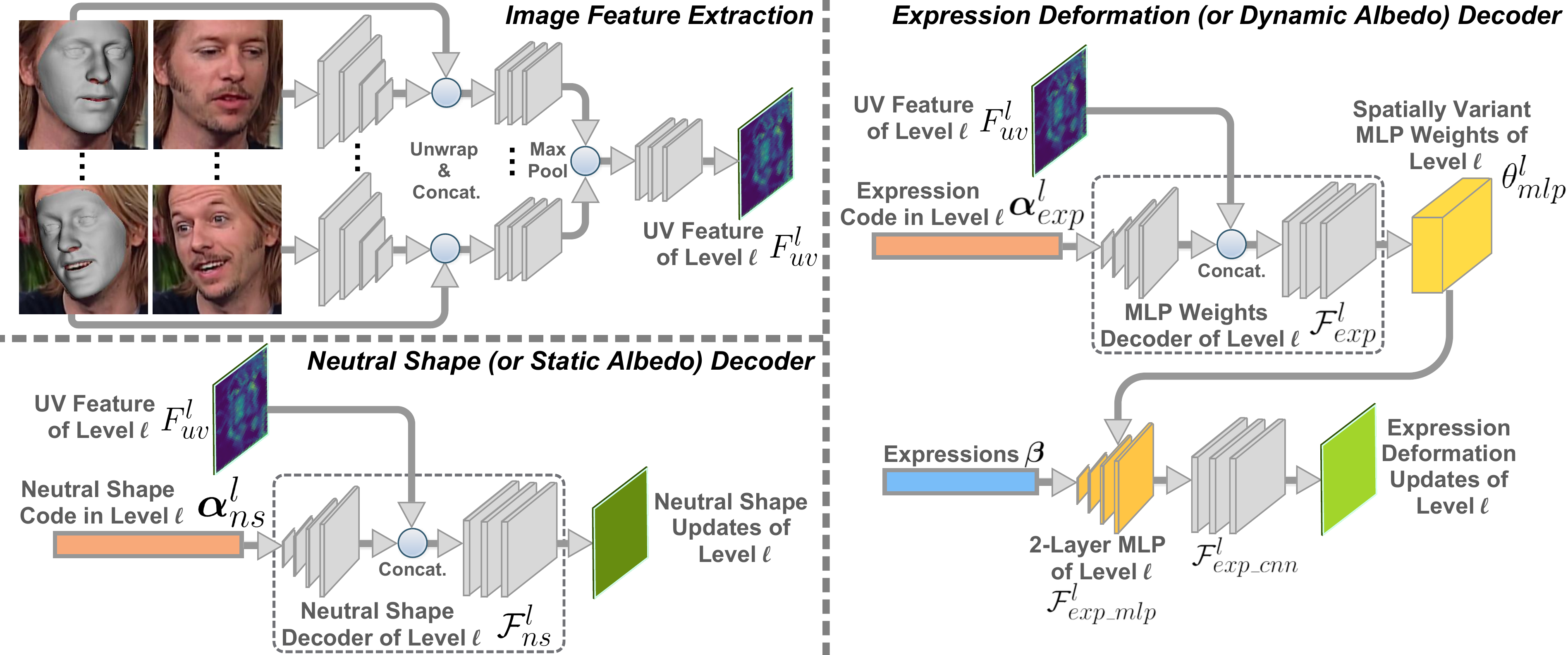}
    \caption{The illustration of different components in our neural rig decoder (\secref{sec:rig_decoder}). Top left: \textbf{\textit{Image Feature Extraction}}. Bottom left: \textbf{\textit{Neutral Shape (or Static Albedo) Decoding}}. Right: \textbf{\textit{Expression Deformation (or Dynamic Albedo) Decoding}}.
    }
	\label{fig:rig_decoder}
	\vspace{-1.25em}
\end{figure*}

\vspace{-.5em}
\subsubsection{Image Feature Extraction}
\vspace{-.25em}
\label{sec:F_uv}
As shown in the top left of \figref{fig:rig_decoder}, given the input images $\{\mathbf{I}_i\}_{i=1}^{N}$ and  initial/intermediate reconstructions $\{\mathbf{\hat{V}}^{old}_i\}_{i=1}^{N}$, we aim to extract a feature map in UV space $F_{uv}$ that encodes the priors derived from visual clues for the later face rig decoding. This process is similar to the adaptive basis generation in~\cite{bai2020deep}, with the output linear basis replaced by $F_{uv}$. We use $N$ Siamese branches to extract $N$ feature maps in UV space from $\{\mathbf{I}_i\}$ and $\{\mathbf{\hat{V}}^{old}_i\}$, followed by max pooling and ResBlocks to get the desired UV feature $F_{uv}$.

Due to the $3$-level scheme, we perform this feature extraction at the beginning of each level $l$ with images $\{\mathbf{I}_i\}$ and reconstructed meshes of previous level $\{\mathbf{\hat{V}}^{l-1}_i\}$, resulting $3$ UV feature maps of increasing resolutions $\{F_{uv}^l\}_{l=1}^3$.

\vspace{-.5em}
\subsubsection{Neutral Shape}
\vspace{-.25em}
As shown in the bottom left of \figref{fig:rig_decoder}, given the UV feature map $F_{uv}$ and the neutral shape code $\bm{\alpha}_{ns}$ that is a portion of the rig code $\bm{\alpha}$, we aim to compute a neutral shape $\mathbf{V}_{ns}$ which is a $3$D face mesh in neutral expression. A CNN structure $\mathcal{F}_{ns}(.)$ is used to decode $\bm{\alpha}_{ns}$ and $F_{uv}$ into $\mathbf{V}_{ns}$ (or updates of $\mathbf{V}_{ns}$ in the $3$-level scheme), which consists of several ResBlocks (w/o BatchNorm) and upsampling. 

For the $3$-level scheme, we repeat the mentioned decoding process for $3$ times to increase resolutions and sum up the results. At the $1$st level, we also include a PCA model from \textit{Basel Face Model} (BFM) \cite{paysan20093d} to better leverage the statistical prior. Thus, we  formulate the neutral shape as:
\begin{align}
\label{eq:ns_dec}
    \mathbf{V}_{ns} = \overline{\mathbf{V}} + B_{bfm} \bm{\alpha}_{ns}^1 + \sum_{l=1}^3 \mathcal{F}_{ns}^l(\bm{\alpha}_{ns}^l; F_{uv}^l),
\end{align}
where $\overline{\mathbf{V}}$ and $B_{bfm}$ are the mean shape and PCA bases from BFM~\cite{paysan20093d}. Note that the $3$ levels of decoding processes are gradually added into the formulation level-by-level during the multi-level optimization (\secref{sec:solve_opt}).

\vspace{-.5em}
\subsubsection{Expression Deformation}
\vspace{-.25em}
As shown in the right of \figref{fig:rig_decoder}, given the UV feature map $F_{uv}$, the expression code $\bm{\alpha}_{exp}$ that is a portion of the rig code $\bm{\alpha}$, and the expression parameter $\bm{\beta}$, we aim to compute an expression deformation $\mathbf{D}_{exp}$ which are the per-vertex displacements added on the neutral shape. Three sub-networks are used to get $\mathbf{D}_{exp}$ including $\mathcal{F}_{exp}$, $\mathcal{F}_{exp\_mlp}$, and $\mathcal{F}_{exp\_cnn}$. The networks $\mathcal{F}_{exp\_mlp}$ and $\mathcal{F}_{exp\_cnn}$ define a mapping from the expression parameter $\bm{\beta}$ to the final expression deformation $\mathbf{D}_{exp}$ as
\begin{align}
    \mathbf{D}_{exp} = \mathcal{F}_{exp\_cnn}(\mathcal{F}_{exp\_mlp}(\bm{\beta}; \theta_{mlp})),
\end{align}
where $\mathcal{F}_{exp\_mlp}$ is a $2$-layer MLP with spatially variant weights $\theta_{mlp}$ and $\mathcal{F}_{exp\_cnn}$ is a CNN.
Then, $\mathcal{F}_{exp}$ controls (or personalizes) this mapping by modifying the network weights $\theta_{mlp}$ according to the expression code $\bm{\alpha}_{exp}$ and the UV feature map $F_{uv}$ as
\begin{align}
    \theta_{mlp} = \mathcal{F}_{exp}(\bm{\alpha}_{exp}; F_{uv}).
\end{align}

For the $3$-level scheme, we repeat the mentioned decoding process for $3$ times in increasing resolutions and sum up the results. Note that to utilize the statistical prior, we adopt the expression PCA bases \cite{deng2019accurate,bai2020deep}, built from Facewarehouse \cite{cao2013facewarehouse}, in the $1$st level, thus resulting in some modifications on the network architecture. More specifically, $\mathcal{F}_{exp\_mlp}^1$ is a $2$-layer MLP with spatially invariant weights $\theta_{mlp}^1$, and $\mathcal{F}_{exp\_cnn}^1$ is replaced by a matrix multiplication with the expression PCA bases $B_{exp}$.
Formally, we have
\begin{align}
\label{eq:exp_dec}
\nonumber
    \mathbf{D}_{exp} &= B_{exp} \mathcal{F}_{exp\_mlp}^1(\bm{\beta}; \theta_{mlp}^1) \\ &+ \sum_{l=2}^3 \mathcal{F}_{exp\_cnn}^l(\mathcal{F}_{exp\_mlp}^l(\bm{\beta}; \theta_{mlp}^l)),
\end{align}
where $\theta_{mlp}^l = \mathcal{F}_{exp}^l(\bm{\alpha}_{exp}^l; F_{uv}^l)$, $l = 1, 2, 3$. The final mesh can be obtained by $\mathbf{V} = \mathbf{V}_{ns} + \mathbf{D}_{exp}$. Similar to the neutral shape, the $3$ levels of decoding processes are gradually added into the formulation level-by-level during the multi-level optimization (\secref{sec:solve_opt}).

\vspace{-.5em}
\subsubsection{Albedo}
\vspace{-.25em}
Following \cite{chaudhuri2020personalized}, we also estimate dynamic albedo maps to better capture facial details such as wrinkles. Given the UV feature map $F_{uv}$, the albedo code $\bm{\alpha}_{alb}$, and the expression parameter $\bm{\beta}$, we aim to compute the per-vertex albedo $\mathbf{A}$. Since only a small amount of high-frequency details could vary with expressions, we first estimate a static albedo at the $1$st and $2$nd levels  similar to the neutral shape, then add the dynamic components at the $3$rd level similar to the expression deformation.
Formally, we have
\begin{align}
\label{eq:alb_dec}
\nonumber
    \mathbf{A} &= \overline{\mathbf{A}} + \sum_{l=1}^2 \mathcal{F}_{alb}^l(\bm{\alpha}_{alb}^l; F_{uv}^l) \\ &+ \mathcal{F}_{alb\_cnn}^3(\mathcal{F}_{alb\_mlp}^3(\bm{\beta}; \theta_{mlp}^3)),
\end{align}
where $\theta_{mlp}^3 = \mathcal{F}_{alb}^3(\bm{\alpha}_{alb}^3; F_{uv}^3)$ and $\overline{\mathbf{A}}$ is the average albedo map from BFM \cite{paysan20093d}. The $3$ levels of decoding processes are also gradually added into the formulation level-by-level during the multi-level optimization (\secref{sec:solve_opt}).

\subsection{Learnable Optimization}
\label{sec:opt}
Given the parameterization of the face rig as in \secref{sec:rig_decoder}, the next step is to optimize the rig code $\bm{\alpha}$ and per-image parameters $\{\mathbf{x}_i = (\bm{\beta}_i, \mathbf{p}_i, \bm{\gamma}_i)\}_{i=1}^{N}$ (\ie expressions $\bm{\beta}_i$, poses $\mathbf{p}_i$, and illuminations $\bm{\gamma}_i$) to obtain the final riggable 3D reconstructions as shown in \figref{fig:overview}. The estimation is done by an energy minimization with end-to-end learnable components. We will first introduce how to to get the per-image reconstructions from the parameters $\bm{\alpha}$ and $\{\mathbf{x}_i\}$ (\secref{sec:param_to_recon}), then describe the energy formulation used to optimize the parameters (\secref{sec:energy}), and finally, 
solve the optimization in a multi-level fashion (\secref{sec:solve_opt}).

\vspace{-.5em}
\subsubsection{Per-image Reconstruction from parameters}
\vspace{-.25em}
\label{sec:param_to_recon}
Given the rig code $\bm{\alpha}$ and per-image parameters $\{\mathbf{x}_i\}_{i=1}^N$, we aim to obtain the per-image reconstructions (\ie one colored 3D mesh $(\mathbf{\hat{V}}_i, \mathbf{\hat{C}}_i)$ for each image), on which the objective energy is computed. For each image, we first decode the rig code $\bm{\alpha}$ and the expression parameter $\bm{\beta}_i$ to a mesh with albedo $(\mathbf{V}_i, \mathbf{A}_i)$ by the neural decoder as in \equref{eq:rig_dec}. Then, this mesh is transformed and projected to the image plane by the weak perspective camera model with pose $\mathbf{p}_i = (s, \mathbf{R}, \mathbf{t})$ (\ie scale $s$, rotation $\mathbf{R} \in SO(3)$, and 2D translation $\mathbf{t} \in \mathbb{R}^2$) as $\mathbf{\hat{V}}_i = s \mathbf{R} \mathbf{V}_i + \mathbf{t}$ and $\mathbf{\Pi}(\mathbf{\hat{V}}_i) = \begin{bmatrix}
    1 & 0 & 0 \\
    0 & 1 & 0 \\
\end{bmatrix} \mathbf{\hat{V}}_i$, where $\mathbf{\Pi}(\cdot)$ is the projection function. Following \cite{tewari2019fml,shang2020self}, we assume Lambertian surface and adopt the Spherical Harmonics (SH) illumination model~\cite{ramamoorthi2001signal} as $\mathbf{\hat{c}}_i = \mathbf{a}_i \cdot \sum_{b=1}^{9} \bm{\gamma}_{i,b} H_b$ to obtain the final mesh color $\mathbf{\hat{C}}_i$, where $\mathbf{\hat{c}}_i$/$\mathbf{a}_i$ is the per-vertex color/albedo.

\vspace{-.5em}
\subsubsection{Energy Formulation}
\vspace{-.25em}
\label{sec:energy}
We define the objective to measure how well the reconstructions $\{(\mathbf{\hat{V}}_i, \mathbf{\hat{C}}_i)\}_{i=1}^N$ explain the input images $\{\mathbf{I}_i\}_{i=1}^N$: 
\begin{align}
    \mathbf{E}(\bm{\alpha}^l, \{\mathbf{x}_i\}) = \lambda_{app} \mathbf{E}_{app} + \lambda_{lan} \mathbf{E}_{lan} + \lambda_{pho} \mathbf{E}_{pho},
\end{align}
with multi-view appearance consistency $\mathbf{E}_{app}$, landmark alignment $\mathbf{E}_{lan}$, and photo-metric reconstruction $\mathbf{E}_{pho}$.

For the multi-view appearance consistency, we follow the formulation of Bai \etal \cite{bai2020deep} and define this term in a feature-metric manner. For each image $\mathbf{I}_i$, we project the reconstructed mesh $\mathbf{\hat{V}}_i$ onto the feature map of $\mathbf{I}_i$ extracted by a FPN \cite{lin2017feature} and do sampling via bilinear interpolation to get per-vertex feature vectors $F(\mathbf{\hat{v}}_i^k)$, where $\mathbf{\hat{v}}_i^k$ denotes the the $k$-th vertex of the mesh $\mathbf{\hat{V}}_i$. We then compute the $L_2$ differences of feature vectors between the corresponding vertices of pairs of images. Formally, we have
\begin{align}
    \mathbf{E}_{app} = \frac{2}{N(N-1)} \sum_{i \neq j} \frac{1}{M} \sum_{k=1}^{M} \|F(\mathbf{\hat{v}}_i^k) - F(\mathbf{\hat{v}}_j^k)\|_2^2,
\end{align}
where $M$ is the number of vertices per mesh. We exclude invisible vertices according to the z-buffer in rendering.

For landmark alignment, we use a similar formulation as \cite{tewari2019fml,tewari2018self} 
with sliding landmarks on the contour, reads as
\begin{align}
\label{eq:lan_obj}
    \mathbf{E}_{lan} = \frac{1}{N} \sum_{i=1}^{N} \frac{1}{68} \sum_{k=1}^{68} \|\mathbf{u}_i^k - \mathbf{\Pi}(\mathbf{\hat{v}}_i^{\mathbf{u}^k})\|_2^2,
\end{align}
where $\mathbf{\hat{v}}_i^{\mathbf{u}^k}$ denotes the mesh vertex corresponding to the $k$-th landmark of image $\mathbf{I}_i$ and $\{\mathbf{u}_i^k\}$ are computed with the state-of-the-art face landmark detector \cite{bulat2017far}.

For the photo-metric reconstruction, we first sample the per-vertex colors $\mathbf{c}_i^k$ from the image $\mathbf{I}_i$ in a similar way for computing the per-vertex feature $F(\mathbf{\hat{v}}_i^k)$. Then, we measure the $L_2$ consistency between the sampled image colors $\mathbf{c}_i^k$ and the reconstructed colors $\mathbf{\hat{c}}_i^k$ from $\mathbf{\hat{C}}_i$. Formally, we have
\begin{align}
    \mathbf{E}_{pho} = \frac{1}{N} \sum_{i=1}^N \frac{1}{M} \sum_{k=1}^M \|\mathbf{c}_i^k - \mathbf{\hat{c}}_i^k\|_2^2,
\end{align}
where visibility is handled in the same way as for $\mathbf{E}_{app}$.

\vspace{-.5em}
\subsubsection{Solving Optimization}
\vspace{-.25em}
\label{sec:solve_opt}
\subtitle{Initialization}
At the beginning, an initialization is needed as the starting of optimization. We obtain the initial face rig by removing all levels of rig decoders in Equs.~\eqref{eq:ns_dec}\eqref{eq:exp_dec}\eqref{eq:alb_dec}, resulting in a constant rig $(\overline{\mathbf{V}}, \overline{\mathbf{A}}) = Rig(\cdot)$. Thus, the decoded mesh with albedo is $(\mathbf{V}_i^0, \mathbf{A}_i^0) = (\overline{\mathbf{V}}, \overline{\mathbf{A}})$ for each image. The initial pose $\mathbf{p}_i^0$ is regressed by a pre-trained network as in \cite{bai2020deep}, which is used to get the per-image mesh $\mathbf{\hat{V}}_i^0$ by transforming $\mathbf{V}_i^0$. Finally, the initial illumination is estimated by solving a linear weighted least square problem with a modified version of $\mathbf{E}_{pho}$ as the objective defined as
\begin{align}
    \mathbf{E}_{pho}^{\prime} = \frac{1}{N} \sum_{i=1}^N \frac{1}{M} \sum_{k=1}^M w_i^k \|\mathbf{c}_i^k - \mathbf{\hat{c}}_i^k\|_2^2,
    \label{equ:illumination}
\end{align}
where $w_i^k$ is a constant set as $(\mathbf{a}_i^k)^{-0.5} \|\mathbf{c}_i^k - \mathbf{\hat{c}}_i^k\|_2^{-1.5}$ and $\mathbf{a}_i^k$ is the per-vertex albedo
to gain better robustness to outliers. For convenience, we treat this initialization as level $0$.

\subtitle{Multi-level Scheme}
For each level $l \geq 1$, we have $3$ steps to update the face rig and per-image parameters. First, given the images $\{\mathbf{I}_i\}$ and the per-image meshes of the previous level $\{\mathbf{\hat{V}}_i^{l-1}\}$, we extract the level $l$ UV feature $F_{uv}^l$ (\secref{sec:F_uv}). Then the level $l$ rig decoders in Equs.~ \eqref{eq:ns_dec}\eqref{eq:exp_dec}\eqref{eq:alb_dec} are added into the rig formulation to parameterize the face rig updates of level $l$ into $\bm{\alpha}^l = (\bm{\alpha}_{ns}^l, \bm{\alpha}_{exp}^l, \bm{\alpha}_{alb}^l)$. $\bm{\alpha}^l$ is set to zeros and $\{\mathbf{x}_i\}$ are set to the outputs of level $l-1$. Second, we solve the optimization $\mathop{\arg\min}_{\bm{\alpha}^l, \{\bm{\beta}_i, \mathbf{p}_i\}} \mathbf{E}$ via gradient descent with step sizes regressed by a network as in \cite{bai2020deep}. Finally, the illumination $\bm{\gamma}_i$ is updated according to Equ.~\eqref{equ:illumination} as the initialization.

\begin{figure}[t!]
    \centering
    \renewcommand{\arraystretch}{0.5}
    \resizebox{1.025\linewidth}{!}{
    \begin{tabular}{@{\hskip -1.5pt}c@{\hskip 1.5pt}c@{\hskip 1.5pt}c@{\hskip 1.5pt}c@{\hskip 1.5pt}c@{\hskip 1.5pt}c@{\hskip 1.5pt}c}
     & Input & Recon. & Shape & Albedo & Lighting & $2^{nd}$ image \\
    \rotatebox[origin=l]{90}{\small Chaudhuri} &
    \includegraphics[height=1.5cm]{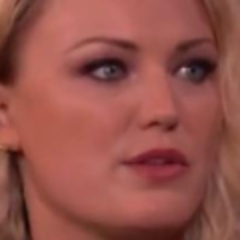} &
    \includegraphics[height=1.5cm]{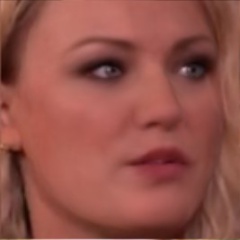} &
    \includegraphics[height=1.5cm]{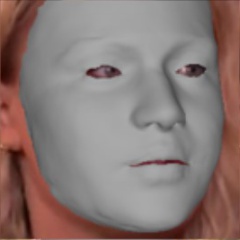} &
    \includegraphics[height=1.5cm]{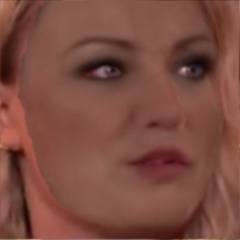} &
    \includegraphics[height=1.5cm]{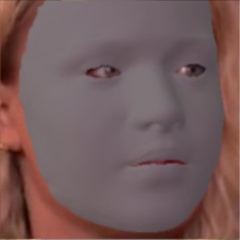} & \\
    \rotatebox[origin=l]{90}{\small Ours} &
    \includegraphics[height=1.5cm]{figs/Chaudhuri20_comp/Ours_case0_input.jpg} &
    \includegraphics[height=1.5cm]{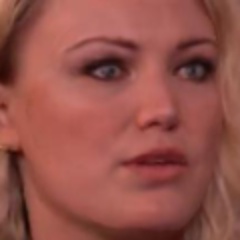} &
    \includegraphics[height=1.5cm]{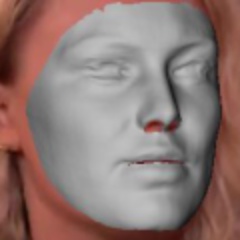} &
    \includegraphics[height=1.5cm]{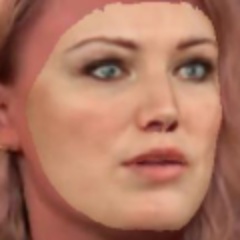} &
    \includegraphics[height=1.5cm]{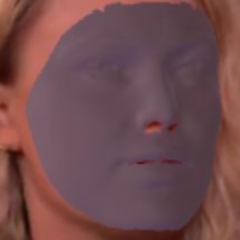} &
    \includegraphics[height=1.5cm]{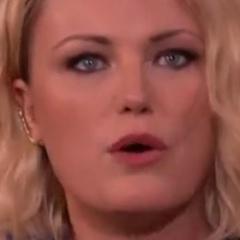} \\
    \rotatebox[origin=l]{90}{\small Chaudhuri} &
    \includegraphics[height=1.5cm]{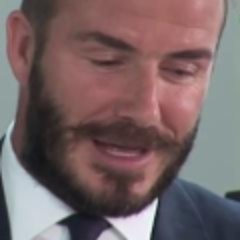} &
    \includegraphics[height=1.5cm]{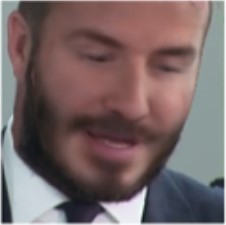} &
    \includegraphics[height=1.5cm]{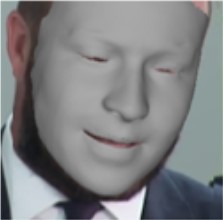} &
    \includegraphics[height=1.5cm]{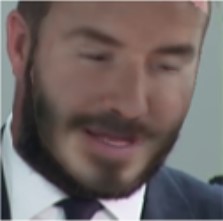} &
    \includegraphics[height=1.5cm]{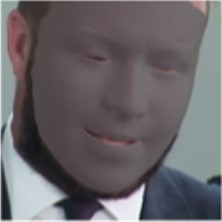} & \\
    \rotatebox[origin=l]{90}{\small Ours} &
    \includegraphics[height=1.5cm]{figs/Chaudhuri20_comp/Ours_case2_input.jpg} &
    \includegraphics[height=1.5cm]{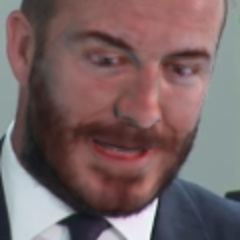} &
    \includegraphics[height=1.5cm]{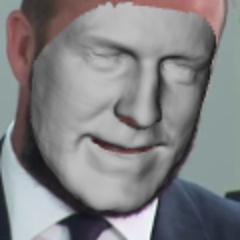} &
    \includegraphics[height=1.5cm]{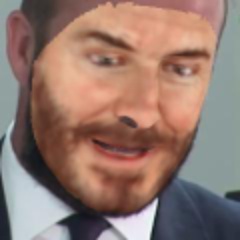} &
    \includegraphics[height=1.5cm]{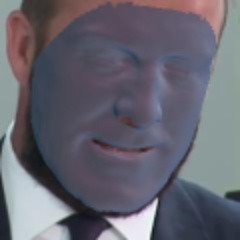} &
    \includegraphics[height=1.5cm]{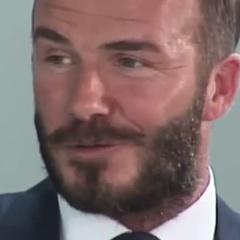} \\
    \end{tabular}
    }

    \caption{Qualitative comparison with Chaudhuri \etal \cite{chaudhuri2020personalized}. Our estimated geometries are superior and textures are comparable. 
    }
    \label{fig:qual_Chaudhuri}
\vspace{-1.em}
\end{figure}

\subsection{Training Losses}
\label{sec:loss}
Our model is trained with registered ground truth $3$D scans in a supervised manner, with ground truth in $3$D scans in the camera space, identity label, and $3$D scans in neutral expression for each identity. The complete loss is
\begin{align}
\nonumber
    L &= L_{pose} + L_{recon\_geo} + L_{ns\_geo} \\
    &+ \lambda_1 L_{recon\_pho} + \lambda_2 L_{\bm{\beta}} + \lambda_3 L_{ns\_con}.
\end{align}




$L_{pose}$ and $L_{recon\_geo}$ are losses supervising the per-image meshes, each of which contains $2$ terms as $L_{pose} = 0.025 L_{lan} + L_{dep\_v}$ and $L_{recon\_geo} = L_{den\_v} + 1000 L_{norm}$. We define these $4$ terms following \cite{bai2020deep}. \ziqian{Please see the supplementary for detailed loss definitions.}

$L_{ns\_geo}$ is a geometry loss supervising the neutral shape reconstruction. It has the same formulation as $L_{recon\_geo}$, except being computed between the ground truth neutral shape and the estimated one.

$L_{recon\_pho}$ is a photo-metric loss supervising the per-image reconstructions. Following \cite{chaudhuri2020personalized}, we use the differentiable rendering to obtain the reconstructed image, then compute the image intensity loss $L_{img}$ and image gradient loss $L_{grad}$ in $L_{2,1}$ norm \cite{towards-high-fidelity-nonlinear-3d-face-morphable-model,chaudhuri2020personalized}. 
These two losses are added together as $L_{recon\_pho} = L_{img} + L_{grad}$.

$L_{\bm{\beta}}$ is an $L_2$ loss to encourage a small expression parameter $\bm{\beta}_i$ when the per-image reconstruction should be in neutral expression. Formally, we have $L_{\bm{\beta}} = \|\bm{\beta}_i\|_2^2$ when the ground truth $\mathbf{V}_i^{gt}$ is in neutral expression.

$L_{ns\_con}$, termed as neutral shape consistency loss, is computed on the estimated neutral shapes to encourage small intra-identity differences. For two estimated neutral shapes of the same identity in a mini-batch, we compute the $L_2$ vertex loss $L_{ns\_con\_v}$ (same form as $L_{dep\_v}$) and the cosine normal loss $L_{ns\_con\_n}$ (same form as $L_{norm}$) between them to enforce the consistency, which reads as $L_{ns\_con} = L_{ns\_con\_v} + 1000L_{ns\_con\_n}$.

\begin{figure}[t!]
    \centering
    \renewcommand{\arraystretch}{0.5}
    \resizebox{1.025\linewidth}{!}{
    \begin{tabular}{@{\hskip -1.5pt}c@{\hskip 1.5pt}c@{\hskip 1.5pt}c@{\hskip 1.5pt}c@{\hskip 1.5pt}c@{\hskip 1.5pt}c@{\hskip 1.5pt}c}
     & Input & Recon. & Albedo & Lighting & Shape & Bai20 \\
    \rotatebox[origin=l]{90}{\small Tewari19} &
    \includegraphics[height=1.5cm]{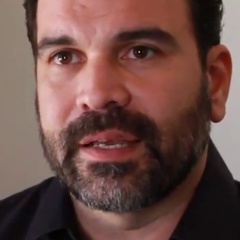} &
    \includegraphics[height=1.5cm]{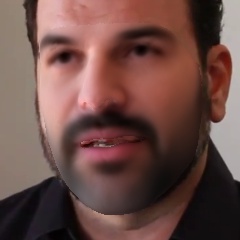} &
    \includegraphics[height=1.5cm]{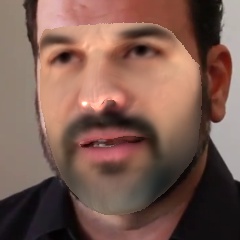} &
    \includegraphics[height=1.5cm]{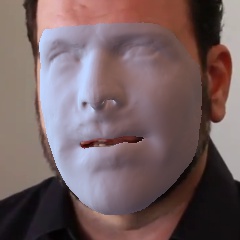} &
    \includegraphics[height=1.5cm]{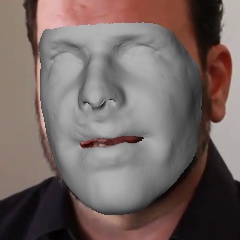} &
    \multirow{2}{*}[3ex]{\includegraphics[height=1.5cm]{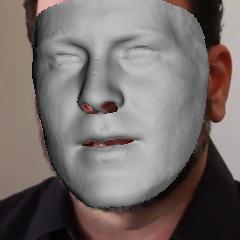}} \\
    \rotatebox[origin=l]{90}{\small Ours} &
    \includegraphics[height=1.5cm]{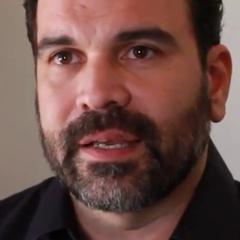} &
    \includegraphics[height=1.5cm]{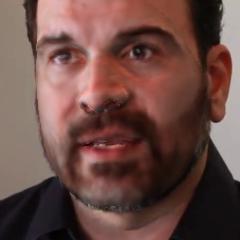} &
    \includegraphics[height=1.5cm]{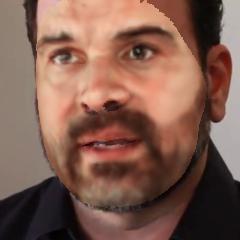} &
    \includegraphics[height=1.5cm]{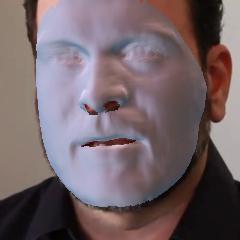} &
    \includegraphics[height=1.5cm]{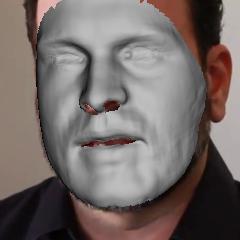} & \\
    \rotatebox[origin=l]{90}{\small Tewari19} &
    \includegraphics[height=1.5cm]{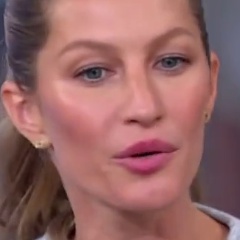} &
    \includegraphics[height=1.5cm]{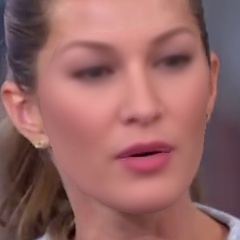} &
    \includegraphics[height=1.5cm]{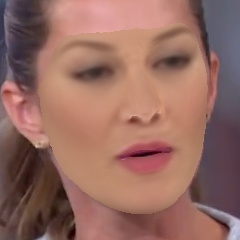} &
    \includegraphics[height=1.5cm]{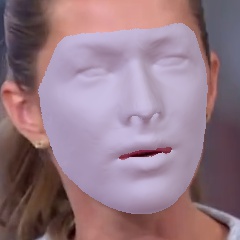} &
    \includegraphics[height=1.5cm]{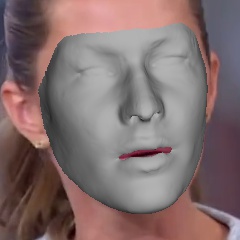} &
    \multirow{2}{*}[3ex]{\includegraphics[height=1.5cm]{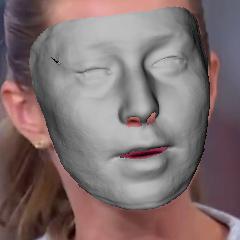}} \\
    \rotatebox[origin=l]{90}{\small Ours} &
    \includegraphics[height=1.5cm]{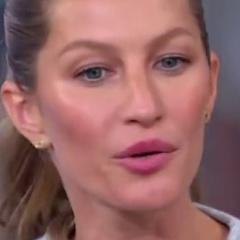} &
    \includegraphics[height=1.5cm]{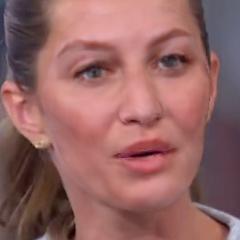} &
    \includegraphics[height=1.5cm]{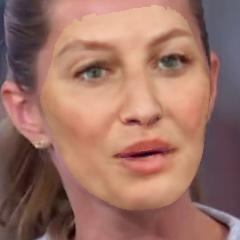} &
    \includegraphics[height=1.5cm]{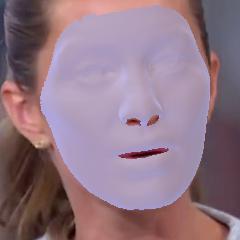} &
    \includegraphics[height=1.5cm]{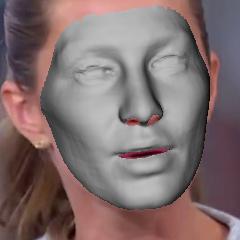} & \\
    \end{tabular}
    }

    \caption{Qualitative comparison with Tewari~\etal~\cite{tewari2019fml} and Bai~\etal~\cite{bai2020deep}. We obtain more faithful geometries that better reflect the corresponding personnel and higher resolution textures. 
    }
    \label{fig:qual_bai_fml}
\vspace{-1em}
\end{figure}

\begin{table*}[!t]
\begin{tabular}{ccc}

\begin{minipage}{0.385\textwidth}
    \setlength\tabcolsep{4pt}
    \resizebox{0.9\width}{!}{
    \begin{tabular}{c|c|cc|cc}
        \multicolumn{2}{c|}{} & \multicolumn{2}{c|}{Bosphorus} & \multicolumn{2}{c}{BU3DFE} \\
        \multicolumn{2}{c|}{} & Mean & STD & Mean & STD \\
        \hline
        \multicolumn{6}{c}{Protocol of Bai \etal \cite{bai2020deep}} \\
        \hline
        \textit{Single} &
        Tewari18 \cite{tewari2018self} & - & - & 1.78 & 0.49 \\
        \textit{View} & Deng19 \cite{deng2019accurate} & 1.47 & 0.40 & 1.38 & 0.37 \\
        \hline
        \multirow{4}{*}{\makecell{\textit{Two} \\ \textit{Views}}} &
        Tewari19 \cite{tewari2019fml} & - & - & 1.74 & 0.45 \\
        & Bai20 \cite{bai2020deep} & 1.44 & \textbf{0.38} & 1.11 & 0.29 \\
        & Ours(R) & 1.37 & 0.44 & 1.09 & 0.31 \\
        & Ours & \textbf{1.36} & \textbf{0.38} & \textbf{1.00} & \textbf{0.27} \\
    \end{tabular}
    }
\caption{Geometry accuracy on Bosphorus \cite{savran2008bosphorus} and BU3DFE \cite{yin20063d} with testing protocol of Bai \etal \cite{bai2020deep}.}
\label{tab:quan_bai}
\end{minipage}

&

\begin{minipage}{0.31\textwidth}
    \setlength\tabcolsep{4pt}
    \resizebox{0.9\width}{!}{
    \begin{tabular}{c|c|cc}
        \multicolumn{2}{c|}{} & \multicolumn{2}{c}{BU3DFE} \\
        \multicolumn{2}{c|}{} & Mean & STD \\
        \hline
        \multicolumn{4}{c}{Protocol of Tewari \etal \cite{tewari2019fml}} \\
        \hline
        \textit{Single} &
        Tewari18 \cite{tewari2018self} & 1.83 & 0.39 \\
        \textit{View} & Shang20 \cite{shang2020self} & 1.55 & 0.31 \\
        \hline
        \multirow{4}{*}{\makecell{\textit{Two} \\ \textit{Views}}} &
        Tewari19 \cite{tewari2019fml} & 1.78 & 0.45 \\
        & Chaudhuri20 \cite{chaudhuri2020personalized} & 1.61 & 0.32 \\
        & Ours(R) & 1.27 & 0.26 \\
        & Ours & \textbf{1.21} & \textbf{0.25} \\
    \end{tabular}
    }
\caption{Geometry accuracy on BU3DFE \cite{yin20063d} with protocol of Tewari \etal \cite{tewari2019fml}.}
\label{tab:quan_tewari}
\end{minipage}

&

\begin{minipage}{0.24\textwidth}
\setlength\tabcolsep{4pt}
    \resizebox{0.9\width}{!}{
    \begin{tabular}{c|cc}
         & \multicolumn{2}{c}{NoW Dataset} \\
         & Mean & STD \\
        \hline
        \multicolumn{3}{c}{\textit{Single View}} \\
        \hline
        Tuan17 \cite{tuan2017regressing} & 2.31 & 0.42 \\
        Feng18 \cite{feng2018joint} & 1.97 & 0.68 \\
        Sanyal19 \cite{sanyal2019learning} & 1.52 & 0.50 \\
        \hline
        \multicolumn{3}{c}{\textit{Multi Views}} \\
        \hline
        Ours &\textbf{1.33} & \textbf{0.28} \\
    \end{tabular}
    }
\caption{Geometry accuracy of neutral shapes on NoW~\cite{sanyal2019learning}.}
\label{tab:ns_quan}
\end{minipage}

\end{tabular}
\vspace{-1.5em}
\end{table*}

\section{Experiments}
\label{sec:exp_setup}
\subtitle{Training Data}
Our model is trained on Stirling/ESRC 3D face database \cite{esrc}, containing textured 3D scans of $100+$ subjects, each with up to $8$ expressions. We synthesize training data by rendering 3D scans of $116$ subjects ($31$ for validation). 
For each training/validation sample, we first randomly sample $10$ expressive scans with replacement (\ie can have repeated expressions)
of the same identity,
then each scan is associated with a random pose and an illumination sampled from the training data of Sengupta~\etal~\cite{sengupta2018sfsnet}. Finally, we use the selected scans, poses, and illuminations to render $10$ images as a training/validation sample. 
Our training losses require dense vertex correspondences between the reconstructions and scans (\ie registered scans). Following~\cite{bai2020deep}, we first fit the 3DMM (BFM + expression bases) to the landmarks of each scan, then perform Non-Rigid ICP \cite{amberg2007optimal} to obtain the dense correspondences.

\subtitle{Implementation}
Our algorithm is implemented via Pytorch. To  optimize each level, we perform $3$ iterations of parameter updates with weights $\lambda_{app} = 0.25$, $\lambda_{lan} = 0.025$, and $\lambda_{pho} = 1$. During training, we randomly select $2$-$7$ out of $10$ images for each data sample to make our model adapt to different numbers of images. When computing $\mathbf{E}_{app}$ during training, we randomly sample vertices to save GPU memory. To focus our model on reconstruction quality, we remove the neutral shape consistency loss $L_{ns\_con}$ for now (\ie $\lambda_3 = 0$). The rest loss weights are $\lambda_1 = \lambda_2 = 10$, and the batch size is $1$ with a learning rate $2.0 \times 10^{-5}$.


\subsection{Per-image Reconstruction}
Our method can estimate a 3D face per input image. We evaluate 3D reconstruction quantitatively and qualitatively.

\subtitle{Quantitative Evaluation}
Two datasets BU3DFE \cite{yin20063d} and Bosphorus \cite{savran2008bosphorus} are used to evaluate the 3D reconstructions. Following \cite{bai2020deep}, we first roughly align the predicted mesh to the 3D scan and crop the scan based on landmarks ($8$ for BU3DFE and $5$ for Bosphorus), then perform ICP \cite{Zhou2018} to improve the alignment. Finally, per-vertex point-to-plane distances from 3D scans to reconstructions are computed. On BU3DFE, we also follow the protocol of Tewari \etal \cite{tewari2019fml} to align and compute error based on dense correspondences for fairly comparison with Chaudhuri \etal \cite{chaudhuri2020personalized}, whose numbers are cited from their paper. As in \tabref{tab:quan_bai} and \tabref{tab:quan_tewari}, our method outperforms various single- and multi-image monocular methods, achieving the state of the art.

\subtitle{Qualitative Evaluation}
We also visually compare with previous works on VoxCeleb2 \cite{chung2018voxceleb2} dataset. \figref{fig:qual_Chaudhuri} shows the comparison with Chaudhuri \etal \cite{chaudhuri2020personalized}. In terms of geometry, our method outperforms~\cite{chaudhuri2020personalized} by capturing more medium level details such as smiling lines, better nose shapes, and geometry around eyes. In terms of texture, we obtain comparable results, with more high-frequency details but also slightly more artifacts. Note that our method additionally takes in the second image shown in~\figref{fig:qual_Chaudhuri}. Compared to \cite{bai2020deep} and \cite{tewari2019fml} in \figref{fig:qual_bai_fml}, our estimated shapes are more faithful against the input images and better reflects the corresponding personnel. Also, our estimated textures are in a higher resolution comparing to Tewari \etal \cite{tewari2019fml}, while Bai \etal \cite{bai2020deep} does not estimate textures.
More video and image results can be found in the supplementary material.

\subsection{Neutral Shape Reconstruction}
\label{sec:ns_eval}
One natural property of face rigs is the ability to disentangle neutral shapes and expression deformations. Therefore, we also evaluate the reconstructed neutral shape as a measure of the face rig quality, by using standard geometry accuracy as well as 3D face verification rate.

\subtitle{Geometry Accuracy} We evaluate our neutral shapes on the NoW Dataset \cite{sanyal2019learning}, which contains $2,054$ $2$D images of $100$ subjects and a separate neutral shape $3$D scan for each subject. Under each subject, the images are classified into $4$ categories (\ie neutral, expression, occlusion and selfie).

Since the dataset is originally designed for single-image methods, we use the following protocol to adapt it to our multi-image setting.
For each subject, we run our algorithm separately for images of different categories, resulting in $4$ neutral shapes. The motivation is to make our setting  closer to the original single-image setting.
When only $1$ image is available in some cases, we horizontally flip it to generate the second input image.  Finally, geometry errors between reconstructed neutral shapes and ground truth scans are computed as in \cite{sanyal2019learning} in a per-image manner (\ie one neutral shape is tested as many times as the number of images used to reconstruct it) to be consistent with the single-image setting.
As  in \tabref{tab:ns_quan}, our approach outperforms prior monocular methods. Note that we only compare with single-image methods since we were not able to find multi-image methods that can separate neutral shapes and expressions while having a public implementation. 

\subtitle{3D Face Verification}
Quantitative evaluations in \tabref{tab:quan_bai}, \ref{tab:quan_tewari}, and \ref{tab:ns_quan} are performed on datasets of a small number of subjects with images taken in controlled setting. However, it is important to evaluate how 3D face reconstruction works on in-the-wild images from a large number of subjects, a challenge without ground truth 3D scans. Thus, instead of computing geometry errors, following~\cite{liu2018joint}, we resort to 3D face verification rate to measure the neutral shape quality.

To this end, we test on the Labeled Faces in the Wild (LFW) \cite{huang2008labeled} benchmark. We estimate the neutral shape of each image via the tested method, then train a network to perform 3D face verification on the estimated shapes. Our method inputs the original image and its flipped one. 
\ziqian{Please see supplementary for details of the verification setup.}

To increase the robustness and discriminativeness, we finetune our model with $\lambda_3 = 5$ and augment images with synthetic degradation.
Limitated by GPU memory, we only select $2$ out of $10$ images per sample and set batch size to $2$.
We denote this version of model as \textit{Ours(R)}. 

\begin{figure}[t!]
\centering
\begin{tabular}{@{\hskip 1pt}c@{\hskip 8pt}c}
\centering
\begin{minipage}{0.255\textwidth}
    \centering
    \includegraphics[width=0.9\linewidth]{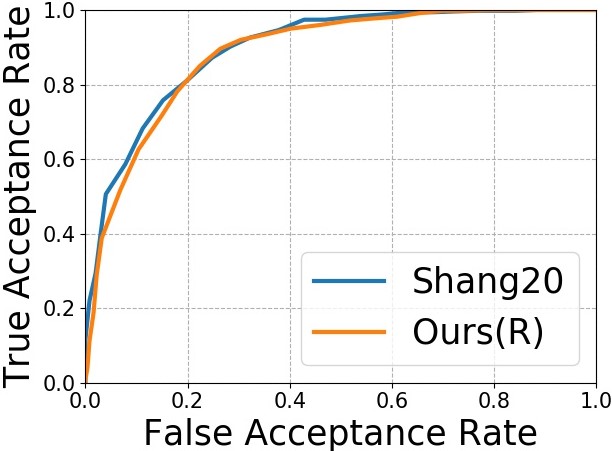}
    \caption{$3$D face verification on LFW~\cite{huang2008labeled}.}
    \label{fig:veri}
\end{minipage}
&
\begin{minipage}{0.2\textwidth}
    \centering
    \setlength\tabcolsep{2.5pt}
    \begin{tabular}{c|cc}
         & Mean & STD \\
        \hline
        \makecell{W/o $\bm{\alpha}_{exp}$ \\ \& $\bm{\alpha}_{ns}$} & 1.48 & 0.41 \\
        \hline
        W/o $\bm{\alpha}_{ns}$ & 1.41 & 0.42 \\
        \hline
        W/o $\bm{\alpha}_{exp}$ & 1.45 & 0.40 \\
        \hline
        Full & \textbf{1.36} & \textbf{0.38}
    \end{tabular}
    \captionof{table}{Comparing with regression baselines on Bosphorus \cite{savran2008bosphorus}.}
    \label{tab:opt_vs_regr}
\end{minipage}
\end{tabular}
\vspace{-.5em}
\end{figure}

We compare \textit{Ours(R)} to Shang20 \cite{shang2020self}, 
 a self-supervised monocular 3D face reconstruction method.
Note that\cite{shang2020self} is trained on a large amount of in-the-wild faces, serving as a strong baseline on generalization and robustness. From the verification accuracy (Ours(R): $81.4$\%, \cite{shang2020self}: $81.0$\%) and \figref{fig:veri}, two methods perform comparably, demonstrating that even on in-the-wild faces of diverse identities, our method still has reasonable generalization and robustness. 

\subsection{Retargeting}

Retargeting is a standard application of riggable $3$D face reconstruction, where the reconstructed rig of the target actor is re-animated by \ziqian{an image/}video of a source actor. \ziqian{We quantitatively evaluate self-retargeting and qualitatively evaluate cross-actor video retargeting.}

\subtitle{Self-Retargeting}
\ziqian{
Similar to \secref{sec:exp_setup}, we evaluate on synthetic data rendered from $31$ test subjects. We render $284$ samples, each has $8$ images with different expressions. We use $7$ images to build the face rig and use the left one as a novel expression for self-retargeting. Specifically, we run our method on the left image and its flipped version to obtain the expression parameter (\emph{exp param}) of this novel expression. Then we apply \emph{exp param} to the estimated rig and compute geometry errors. We also include reconstruction errors of $7$ viewed expressions as a reference. Since the code of \cite{chaudhuri2020personalized} is not published, we design two baselines based on \cite{shang2020self}: 1) B1: Use PCA bases of \cite{shang2020self} to model the expression space, along with our estimated neutral shape, to form a rig. We obtain \emph{exp param}  by the regressor in \cite{shang2020self}; 2) B2: Replace the neutral shape of rig in B1 with averaged neutral shape obtained from \cite{shang2020self}. As in \tabref{tab:self_retar}, our method has better accuracy than the baselines on novel expressions. Our error on novel expressions is also close to the viewed expressions, indicating good generalization of expressions.}

\begin{table}[!t]
\setlength\tabcolsep{5pt}
\small
\centering
 \begin{tabular}{c|c|c|c|c}
    Exp. Types & Viewed & \multicolumn{3}{c}{Novel} \\ \hline
    Methods & \multicolumn{2}{c|}{Ours} & B1 & B2 \\
    \hline
    Mean & 0.99 & 1.07 & 1.22 & 1.50 \\
    STD & 0.31 & 0.35 & 0.52 & 0.56
 \end{tabular}
 \caption{Geometry errors of novel and viewed expressions in self-retargeting. 
 Key words: (B1) Use PCA bases of \cite{shang2020self} and our estimated neutral shape. (B2) Replace the neutral shape of rig in B1 with averaged neutral shape obtained from \cite{shang2020self}.
 }
 \label{tab:self_retar}
 \vspace{-1.em}
\end{table}

\begin{figure}[t!]
\centering
\renewcommand{\arraystretch}{0.5}
\resizebox{1.0\linewidth}{!}{
\begin{tabular}{@{\hskip -2.5pt}c@{\hskip -4pt}ccccc}
    \rotatebox[origin=l]{90}{\small {\hskip 7pt} Target} &
    \multicolumn{5}{c}{\includegraphics[width=\linewidth]{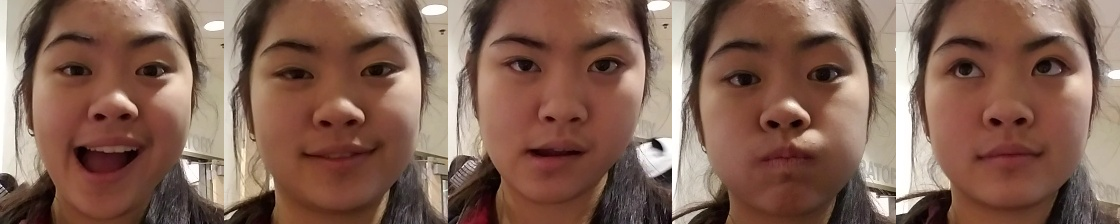}} \\
    \rotatebox[origin=l]{90}{\small Frame 0} & \multicolumn{5}{c}{\includegraphics[width=\linewidth]{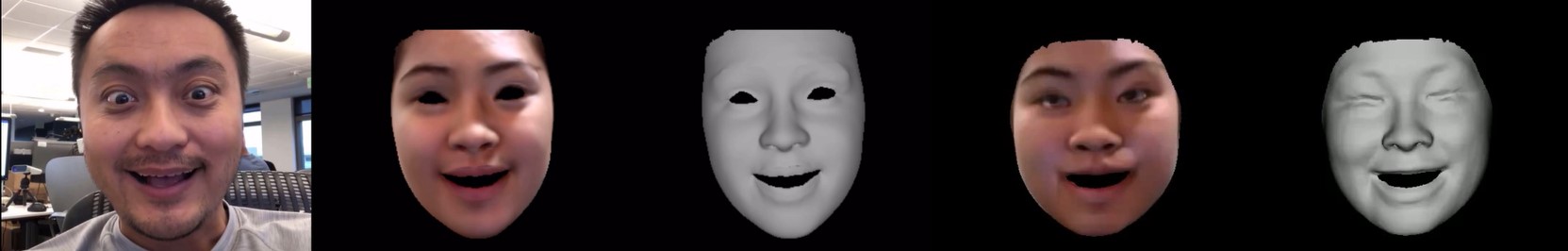}} \\
    \rotatebox[origin=l]{90}{\small Frame 1} & \multicolumn{5}{c}{\includegraphics[width=\linewidth]{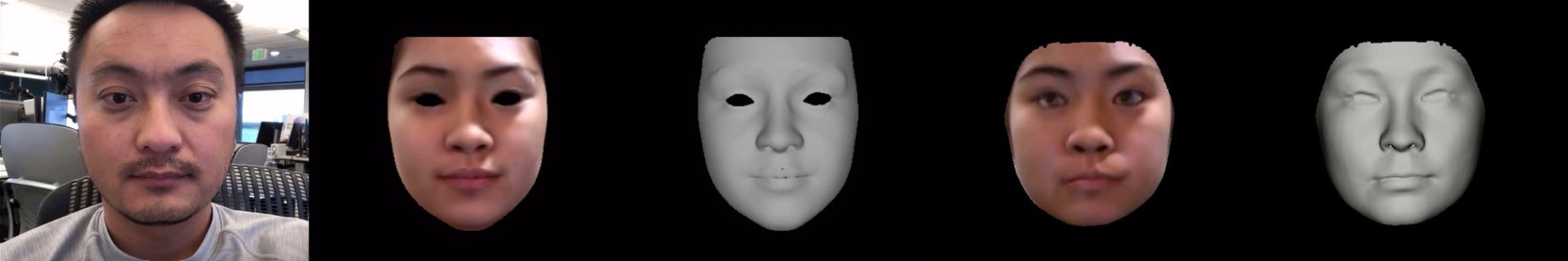}} \\
     & {\hskip 12pt} \small Source & \multicolumn{2}{@{\hskip 35pt}c}{\small Chaudhuri20} & \multicolumn{2}{c}{\small Ours} \\
\end{tabular}
}
\caption{Video retargeting comparing to Chaudhuri \etal \cite{chaudhuri2020personalized}.}
\label{fig:retarget}
\vspace{-1.25em}
\end{figure}

\subtitle{Video Retargeting}
As in \figref{fig:teaser}, our method outputs good results for different targets in both shapes and textures that faithfully reflect the target identity, and reasonably transfers the expressions.
We also visually compare with Chaudhuri \etal \cite{chaudhuri2020personalized} on \ziqian{their demo video.}
Selected 
frames are shown in \figref{fig:retarget}. Our method has superior shape quality better reflecting the personal characteristics, such as the round chin instead of the sharp one from \cite{chaudhuri2020personalized}, and achieves reasonable expression transfer results. Note that in \figref{fig:retarget} we obtain the rig albedo by removing shadings from image colors of the target actor for better visual quality, and the target rig is built by \textit{Ours(R)} for better robustness.
However, our method has a few limitations on unusual expressions, eyelid motion, and the amplitude of transferred expressions. Video results and more analysis can be found in the supplementary.

\subsection{Optimization vs Regression}
The main novelty of our method is the optimizable neural parameterization of the face rig coupled with the learnable optimization. This design introduces optimization into network inference thus explicitly enforces constraints such as multi-view appearance consistency, landmark alignment, and photo-metric reconstruction, which are derived from the first principles based on the domain knowledge. This additional prior information has the potential to improve the 3D reconstruction quality. We investigate this advantage by comparing our method with regression baselines, where the components of the face rig are directly predicted by the neural network instead of being optimized. Please refer to the supplementary for more details of the regression baselines. 
As shown in \tabref{tab:opt_vs_regr}, the performance drops when one or more rig components are regressed, demonstrating the effectiveness of explicit optimization during inference.

\section{Conclusion}
We solve riggable $3$D face reconstruction from monocular RGB images by \ziqian{an end-to-end trainable network embedded with a in-network optimization. The network contains}
an optimisable neural face rig parameterization coupled with \ziqian{a learnable optimization.}
The optimization explicitly enforces first-principal constraints during inference, while the learning components leverage \ziqian{deep}
priors to constrain the \ziqian{ill-posedness}
and alleviate the optimization difficulty. Experiments demonstrate that our method achieves state-of-the-art reconstruction accuracy, reasonable robustness and generalization ability, and can be applied to the standard face rig application such as 
retargeting.

{\small
\bibliographystyle{ieee_fullname}
\bibliography{egbib}

\begin{thebibliography}{10}\itemsep=-1pt

\bibitem{esrc}
{\em Stirling/ESRC 3D face database}.

\bibitem{amberg2007optimal}
Brian Amberg, Sami Romdhani, and Thomas Vetter.
\newblock Optimal step nonrigid icp algorithms for surface registration.
\newblock In {\em IEEE Conf. Comput. Vis. Pattern Recog.}, pages 1--8, 2007.

\bibitem{bai2020deep}
Ziqian Bai, Zhaopeng Cui, Jamal~Ahmed Rahim, Xiaoming Liu, and Ping Tan.
\newblock Deep facial non-rigid multi-view stereo.
\newblock In {\em IEEE Conf. Comput. Vis. Pattern Recog.}, pages 5850--5860,
  2020.

\bibitem{blanz2003face}
Volker Blanz and Thomas Vetter.
\newblock Face recognition based on fitting a 3d morphable model.
\newblock {\em IEEE Trans. Pattern Anal. Mach. Intell.}, 25(9):1063--1074,
  2003.

\bibitem{blanz1999morphable}
Volker Blanz, Thomas Vetter, et~al.
\newblock A morphable model for the synthesis of 3d faces.
\newblock In {\em Proc. of ACM SIGGRAPH}, volume~99, pages 187--194, 1999.

\bibitem{bloesch2018codeslam}
Michael Bloesch, Jan Czarnowski, Ronald Clark, Stefan Leutenegger, and Andrew~J
  Davison.
\newblock Codeslam—learning a compact, optimisable representation for dense
  visual slam.
\newblock In {\em IEEE Conf. Comput. Vis. Pattern Recog.}, pages 2560--2568,
  2018.

\bibitem{bouaziz2013online}
Sofien Bouaziz, Yangang Wang, and Mark Pauly.
\newblock Online modeling for realtime facial animation.
\newblock {\em ACM Trans. Graph.}, 32(4):40, 2013.

\bibitem{bulat2017far}
Adrian Bulat and Georgios Tzimiropoulos.
\newblock How far are we from solving the 2d \& 3d face alignment problem?(and
  a dataset of 230,000 3d facial landmarks).
\newblock In {\em Int. Conf. Comput. Vis.}, pages 1021--1030, 2017.

\bibitem{cao2013facewarehouse}
Chen Cao, Yanlin Weng, Shun Zhou, Yiying Tong, and Kun Zhou.
\newblock Facewarehouse: A 3d facial expression database for visual computing.
\newblock {\em IEEE Trans. Vis. Comput. Graph.}, 20(3):413--425, 2013.

\bibitem{cao2016real}
Chen Cao, Hongzhi Wu, Yanlin Weng, Tianjia Shao, and Kun Zhou.
\newblock Real-time facial animation with image-based dynamic avatars.
\newblock {\em ACM Trans. Graph.}, 35(4), 2016.

\bibitem{chaudhuri2020personalized}
Bindita Chaudhuri, Noranart Vesdapunt, Linda Shapiro, and Baoyuan Wang.
\newblock Personalized face modeling for improved face reconstruction and
  motion retargeting.
\newblock In {\em Eur. Conf. Comput. Vis.}, pages 142--160, 2020.

\bibitem{chung2018voxceleb2}
Joon~Son Chung, Arsha Nagrani, and Andrew Zisserman.
\newblock Voxceleb2: Deep speaker recognition.
\newblock {\em Proc. Interspeech 2018}, pages 1086--1090, 2018.

\bibitem{deng2019accurate}
Yu Deng, Jiaolong Yang, Sicheng Xu, Dong Chen, Yunde Jia, and Xin Tong.
\newblock Accurate 3d face reconstruction with weakly-supervised learning: From
  single image to image set.
\newblock In {\em Proceedings of the IEEE Conference on Computer Vision and
  Pattern Recognition Workshops}, pages 0--0, 2019.

\bibitem{egger20203d}
Bernhard Egger, William~AP Smith, Ayush Tewari, Stefanie Wuhrer, Michael
  Zollhoefer, Thabo Beeler, Florian Bernard, Timo Bolkart, Adam Kortylewski,
  Sami Romdhani, et~al.
\newblock 3d morphable face models—past, present, and future.
\newblock {\em ACM Trans. Graph.}, 39(5):1--38, 2020.

\bibitem{feng2018joint}
Yao Feng, Fan Wu, Xiaohu Shao, Yanfeng Wang, and Xi Zhou.
\newblock Joint 3d face reconstruction and dense alignment with position map
  regression network.
\newblock In {\em Eur. Conf. Comput. Vis.}, pages 534--551, 2018.

\bibitem{garrido2016reconstruction}
Pablo Garrido, Michael Zollh{\"o}fer, Dan Casas, Levi Valgaerts, Kiran
  Varanasi, Patrick P{\'e}rez, and Christian Theobalt.
\newblock Reconstruction of personalized 3d face rigs from monocular video.
\newblock {\em ACM Trans. Graph.}, 35(3):28, 2016.

\bibitem{genova2018unsupervised}
Kyle Genova, Forrester Cole, Aaron Maschinot, Aaron Sarna, Daniel Vlasic, and
  William~T Freeman.
\newblock Unsupervised training for 3d morphable model regression.
\newblock In {\em IEEE Conf. Comput. Vis. Pattern Recog.}, pages 8377--8386,
  2018.

\bibitem{hadsell2006dimensionality}
Raia Hadsell, Sumit Chopra, and Yann LeCun.
\newblock Dimensionality reduction by learning an invariant mapping.
\newblock In {\em IEEE Conf. Comput. Vis. Pattern Recog.}, volume~2, pages
  1735--1742, 2006.

\bibitem{he2016deep}
Kaiming He, Xiangyu Zhang, Shaoqing Ren, and Jian Sun.
\newblock Deep residual learning for image recognition.
\newblock In {\em IEEE Conf. Comput. Vis. Pattern Recog.}, pages 770--778,
  2016.

\bibitem{hsieh2015unconstrained}
Pei-Lun Hsieh, Chongyang Ma, Jihun Yu, and Hao Li.
\newblock Unconstrained realtime facial performance capture.
\newblock In {\em IEEE Conf. Comput. Vis. Pattern Recog.}, pages 1675--1683,
  2015.

\bibitem{hu2017avatar}
Liwen Hu, Shunsuke Saito, Lingyu Wei, Koki Nagano, Jaewoo Seo, Jens Fursund,
  Iman Sadeghi, Carrie Sun, Yen-Chun Chen, and Hao Li.
\newblock Avatar digitization from a single image for real-time rendering.
\newblock {\em ACM Trans. Graph.}, 36(6):1--14, 2017.

\bibitem{huang2008labeled}
Gary~B Huang, Marwan Mattar, Tamara Berg, and Eric Learned-Miller.
\newblock Labeled faces in the wild: A database forstudying face recognition in
  unconstrained environments.
\newblock In {\em Technical Report 07-49, UMass, Amherst}, 2008.

\bibitem{ichim2015dynamic}
Alexandru~Eugen Ichim, Sofien Bouaziz, and Mark Pauly.
\newblock Dynamic 3d avatar creation from hand-held video input.
\newblock {\em ACM Trans. Graph.}, 34(4):1--14, 2015.

\bibitem{lewis2014practice}
John~P Lewis, Ken Anjyo, Taehyun Rhee, Mengjie Zhang, Frederic~H Pighin, and
  Zhigang Deng.
\newblock Practice and theory of blendshape facial models.
\newblock {\em Eurographics (State of the Art Reports)}, 1(8):2, 2014.

\bibitem{li2010example}
Hao Li, Thibaut Weise, and Mark Pauly.
\newblock Example-based facial rigging.
\newblock {\em ACM Trans. Graph.}, 29(4):1--6, 2010.

\bibitem{li2013realtime}
Hao Li, Jihun Yu, Yuting Ye, and Chris Bregler.
\newblock Realtime facial animation with on-the-fly correctives.
\newblock {\em ACM Trans. Graph.}, 32(4):42--1, 2013.

\bibitem{lin2017feature}
Tsung-Yi Lin, Piotr Doll{\'a}r, Ross Girshick, Kaiming He, Bharath Hariharan,
  and Serge Belongie.
\newblock Feature pyramid networks for object detection.
\newblock In {\em IEEE Conf. Comput. Vis. Pattern Recog.}, pages 2117--2125,
  2017.

\bibitem{liu2018joint}
Feng Liu, Qijun Zhao, Xiaoming Liu, and Dan Zeng.
\newblock Joint face alignment and 3d face reconstruction with application to
  face recognition.
\newblock {\em IEEE Trans. Pattern Anal. Mach. Intell.}, 42(3):664--678, 2018.

\bibitem{lombardi2018deep}
Stephen Lombardi, Jason Saragih, Tomas Simon, and Yaser Sheikh.
\newblock Deep appearance models for face rendering.
\newblock {\em ACM Trans. Graph.}, 37(4):1--13, 2018.

\bibitem{paysan20093d}
Pascal Paysan, Reinhard Knothe, Brian Amberg, Sami Romdhani, and Thomas Vetter.
\newblock A 3d face model for pose and illumination invariant face recognition.
\newblock In {\em 2009 Sixth IEEE International Conference on Advanced Video
  and Signal Based Surveillance}, pages 296--301, 2009.

\bibitem{ramamoorthi2001signal}
Ravi Ramamoorthi and Pat Hanrahan.
\newblock A signal-processing framework for inverse rendering.
\newblock In {\em Proceedings of the 28th annual conference on Computer
  graphics and interactive techniques}, pages 117--128, 2001.

\bibitem{richardson20163d}
Elad Richardson, Matan Sela, and Ron Kimmel.
\newblock 3d face reconstruction by learning from synthetic data.
\newblock In {\em 2016 Fourth International Conference on 3D Vision (3DV)},
  pages 460--469. IEEE, 2016.

\bibitem{adaptive-3d-face-reconstruction-from-unconstrained-photo-collections}
Joseph Roth, Yiying Tong, and Xiaoming Liu.
\newblock Adaptive 3d face reconstruction from unconstrained photo collections.
\newblock In {\em IEEE Conf. Comput. Vis. Pattern Recog.}, pages 4197--4206,
  2016.

\bibitem{sanyal2019learning}
Soubhik Sanyal, Timo Bolkart, Haiwen Feng, and Michael~J Black.
\newblock Learning to regress 3d face shape and expression from an image
  without 3d supervision.
\newblock In {\em IEEE Conf. Comput. Vis. Pattern Recog.}, pages 7763--7772,
  2019.

\bibitem{savran2008bosphorus}
Arman Savran, Ne{\c{s}}e Aly{\"u}z, Hamdi Dibeklio{\u{g}}lu, Oya
  {\c{C}}eliktutan, Berk G{\"o}kberk, B{\"u}lent Sankur, and Lale Akarun.
\newblock Bosphorus database for 3d face analysis.
\newblock In {\em European Workshop on Biometrics and Identity Management},
  pages 47--56. Springer, 2008.

\bibitem{sela2017unrestricted}
Matan Sela, Elad Richardson, and Ron Kimmel.
\newblock Unrestricted facial geometry reconstruction using image-to-image
  translation.
\newblock In {\em Int. Conf. Comput. Vis.}, pages 1576--1585, 2017.

\bibitem{sengupta2018sfsnet}
Soumyadip Sengupta, Angjoo Kanazawa, Carlos~D Castillo, and David~W Jacobs.
\newblock Sfsnet: Learning shape, reflectance and illuminance of faces in the
  wild'.
\newblock In {\em IEEE Conf. Comput. Vis. Pattern Recog.}, pages 6296--6305,
  2018.

\bibitem{shang2020self}
Jiaxiang Shang, Tianwei Shen, Shiwei Li, Lei Zhou, Mingmin Zhen, Tian Fang, and
  Long Quan.
\newblock Self-supervised monocular 3d face reconstruction by occlusion-aware
  multi-view geometry consistency.
\newblock In {\em Eur. Conf. Comput. Vis.}, 2020.

\bibitem{tang2019ba}
Chengzhou Tang and Ping Tan.
\newblock Ba-net: Dense bundle adjustment network.
\newblock {\em Int. Conf. Learn. Represent.}, 2019.

\bibitem{tewari2019fml}
Ayush Tewari, Florian Bernard, Pablo Garrido, Gaurav Bharaj, Mohamed Elgharib,
  Hans-Peter Seidel, Patrick P{\'e}rez, Michael Zollhofer, and Christian
  Theobalt.
\newblock Fml: face model learning from videos.
\newblock In {\em IEEE Conf. Comput. Vis. Pattern Recog.}, pages 10812--10822,
  2019.

\bibitem{tewari2018self}
Ayush Tewari, Michael Zollh{\"o}fer, Pablo Garrido, Florian Bernard, Hyeongwoo
  Kim, Patrick P{\'e}rez, and Christian Theobalt.
\newblock Self-supervised multi-level face model learning for monocular
  reconstruction at over 250 hz.
\newblock In {\em IEEE Conf. Comput. Vis. Pattern Recog.}, pages 2549--2559,
  2018.

\bibitem{towards-high-fidelity-nonlinear-3d-face-morphable-model}
Luan Tran, Feng Liu, and Xiaoming Liu.
\newblock Towards high-fidelity nonlinear 3d face morphable model.
\newblock In {\em IEEE Conf. Comput. Vis. Pattern Recog.}, pages 1126--1135,
  2019.

\bibitem{on-learning-3d-face-morphable-model-from-in-the-wild-images}
Luan Tran and Xiaoming Liu.
\newblock On learning 3d face morphable model from in-the-wild images.
\newblock {\em IEEE Trans. Pattern Anal. Mach. Intell.}, 43(1):157--171, 2019.

\bibitem{tuan2017regressing}
Anh Tuan~Tran, Tal Hassner, Iacopo Masi, and G{\'e}rard Medioni.
\newblock Regressing robust and discriminative 3d morphable models with a very
  deep neural network.
\newblock In {\em IEEE Conf. Comput. Vis. Pattern Recog.}, pages 5163--5172,
  2017.

\bibitem{vlasic2006face}
Daniel Vlasic, Matthew Brand, Hanspeter Pfister, and Jovan Popovic.
\newblock Face transfer with multilinear models.
\newblock In {\em ACM SIGGRAPH 2006 Courses}, pages 24--es. 2006.

\bibitem{wu2016anatomically}
Chenglei Wu, Derek Bradley, Markus Gross, and Thabo Beeler.
\newblock An anatomically-constrained local deformation model for monocular
  face capture.
\newblock {\em ACM Trans. Graph.}, 35(4):1--12, 2016.

\bibitem{wu2018deep}
Chenglei Wu, Takaaki Shiratori, and Yaser Sheikh.
\newblock Deep incremental learning for efficient high-fidelity face tracking.
\newblock {\em ACM Trans. Graph.}, 37(6):1--12, 2018.

\bibitem{yang2020facescape}
Haotian Yang, Hao Zhu, Yanru Wang, Mingkai Huang, Qiu Shen, Ruigang Yang, and
  Xun Cao.
\newblock Facescape: a large-scale high quality 3d face dataset and detailed
  riggable 3d face prediction.
\newblock In {\em IEEE Conf. Comput. Vis. Pattern Recog.}, pages 601--610,
  2020.

\bibitem{yin20063d}
Lijun Yin, Xiaozhou Wei, Yi Sun, Jun Wang, and Matthew~J Rosato.
\newblock A 3d facial expression database for facial behavior research.
\newblock In {\em 7th international conference on automatic face and gesture
  recognition (FGR06)}, pages 211--216. IEEE, 2006.

\bibitem{Zhou2018}
Qian-Yi Zhou, Jaesik Park, and Vladlen Koltun.
\newblock {Open3D}: {A} modern library for {3D} data processing.
\newblock {\em arXiv:1801.09847}, 2018.

\bibitem{zollhofer2018state}
Michael Zollh{\"o}fer, Justus Thies, Pablo Garrido, Derek Bradley, Thabo
  Beeler, Patrick P{\'e}rez, Marc Stamminger, Matthias Nie{\ss}ner, and
  Christian Theobalt.
\newblock State of the art on monocular 3d face reconstruction, tracking, and
  applications.
\newblock In {\em Comput. Graph. Forum}, volume~37, pages 523--550, 2018.

\end{thebibliography}
}

\newpage
\renewcommand\thesection{\Alph{section}}
\renewcommand\thetable{\Alph{table}}
\renewcommand\thefigure{\Alph{figure}}

\setcounter{section}{0}
\setcounter{figure}{0}
\setcounter{table}{0}
\setcounter{equation}{0}

To make our paper self-contained, more information is provided in this supplementary material, including more method details (\secref{sec:method}), training/testing data consistency (\secref{sec:data_consist}), additional qualitative results (\secref{sec:qual}), details of the $3$D face verification (\secref{sec:veri}) and the regression baselines (\secref{sec:regr}), quantitative photometric errors (\secref{sec:photo_err}) and limitation analysis (\secref{sec:limit}).

\section{Method Details}
\label{sec:method}
In this section, we provide additional details of various components in our method. It is better to read together with the corresponding sections of the main paper.

\subsection{Image Feature Extraction (Main Paper Sec. 3.1.1)}
We use the following strategy to compute the UV space feature $F_{uv}$ from the input images $\{\mathbf{I}_i\}_{i=1}^{N}$ and  initial/intermediate reconstructions $\{\mathbf{\hat{V}}^{old}_i\}_{i=1}^{N}$. For each image $\mathbf{I}_i$, a feature map is firstly computed by a \textit{Feature Pyramid Network} (FPN) \cite{lin2017feature}. Then, we unwrap the feature map into UV space based on the initial/intermediate reconstruction $\mathbf{\hat{V}}^{old}_i$. This feature map is then concatenated with the $3$-channel UV image that stores the xyz coordinates of $\mathbf{\hat{V}}^{old}_i$, and goes through several ResBlocks \cite{he2016deep}. Finally, we forward the $N$ resulting feature maps into max pooling and ResBlocks to get the desired UV feature $F_{uv}$. Note that we extract $3$ different features $F_{uv}$ for neutral shape, expression deformation, and albedo separately.

\subsection{Neutral Shape (Main Paper Sec. 3.1.2)}
The neural network $\mathcal{F}_{ns}$ is used to compute the neutral shape $\mathbf{V}_{ns}$ from the UV feature map $F_{uv}$ and the neutral shape code $\bm{\alpha}_{ns}$. More specifically, we first decode the vector $\bm{\alpha}_{ns}$ into a feature map via a FC-layer and several ResBlocks interleaved nearest upsampling, then the resulting feature map is concatenated with $F_{uv}$ to be further decoded into the neutral shape $\mathbf{V}_{ns}$ via ResBlocks.

\subsection{Expression Deformation (Main Paper Sec. 3.1.3)}
As described in Sec. 3.1.3 of the main paper, $3$ sub-networks ($\mathcal{F}_{exp}$, $\mathcal{F}_{exp\_mlp}$, and $\mathcal{F}_{exp\_cnn}$) are used to compute the expression deformation $\mathbf{D}_{exp}$ from the UV feature map $F_{uv}$, the expression code $\bm{\alpha}_{exp}$, and the expression parameter $\bm{\beta}$. More specifically, $\mathcal{F}_{exp}$ is a CNN structure similar to $\mathcal{F}_{ns}$ that decodes $\bm{\alpha}_{exp}$ and $F_{uv}$ into a tensor $\theta_{mlp} \in \mathbb{R}^{H \times W \times (C_{\bm{\beta}} \times C_0 + C_0 \times C_1)}$ (\ie the spatially variant weights of the $2$-layer MLP $\mathcal{F}_{exp\_mlp}$), where $H$ and $W$ are the spatial dimensions of the UV space while $\{C_{\bm{\beta}}, C_0, C_1\}$ are the channel sizes of the expression parameter $\bm{\beta}$, the hidden layer of $\mathcal{F}_{exp\_mlp}$, and the output of $\mathcal{F}_{exp\_mlp}$ respectively. Then the $2$-layer MLP $\mathcal{F}_{exp\_mlp}$ with spatially variant weights $\theta_{mlp}$ decodes the expression parameter $\bm{\beta} \in \mathbb{R}^{C_{\bm{\beta}}}$ to a feature map with size $H \times W \times C_1$, which is further decoded by the CNN $\mathcal{F}_{exp\_cnn}$ to the final expression deformation $\mathbf{D}_{exp}$. Note that we do not personalize the weights of $\mathcal{F}_{exp\_cnn}$ in order not to exceed the memory limitation.

For level $1$ in the $3$-level scheme, we first convert the UV feature $F_{uv}^1$ into a vector with several convolution blocks, then concatenate the vector with the expression code $\bm{\alpha}_{exp}^1$, and feed the concatenated vector into a MLP to obtain the conventional (\ie spatially invariant) weights $\theta_{mlp}^1 \in \mathbb{R}^{C_{\bm{\beta}} \times C_0 + C_0 \times C_1}$.

\begin{figure}[t!]
    \centering
    \renewcommand{\arraystretch}{0.5}
    \resizebox{1.025\linewidth}{!}{
    \begin{tabular}{@{\hskip -1.5pt}c@{\hskip 1.5pt}c@{\hskip 1.5pt}c@{\hskip 1.5pt}c@{\hskip 1.5pt}c@{\hskip 1.5pt}c@{\hskip 1.5pt}c}
     & Input & Recon. & Shape & Albedo & Lighting & $2^{nd}$ image \\

    \rotatebox[origin=l]{90}{\small Chaudhuri} &
    \includegraphics[height=1.5cm]{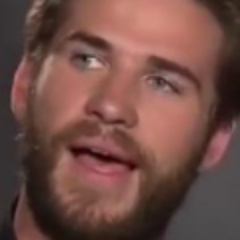} &
    \includegraphics[height=1.5cm]{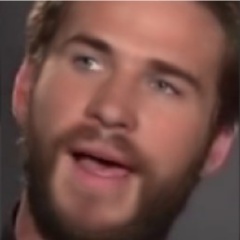} &
    \includegraphics[height=1.5cm]{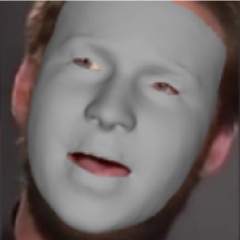} &
    \includegraphics[height=1.5cm]{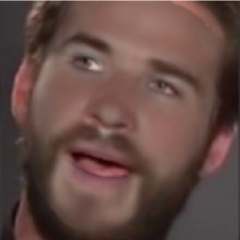} &
    \includegraphics[height=1.5cm]{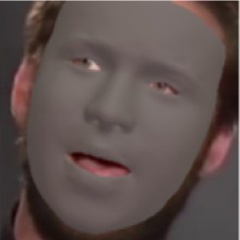} & \\
    \rotatebox[origin=l]{90}{\small Ours} &
    \includegraphics[height=1.5cm]{figs/Chaudhuri20_comp/Ours_case1_input.jpg} &
    \includegraphics[height=1.5cm]{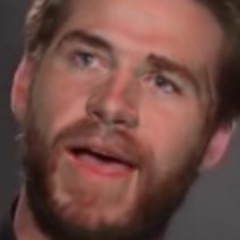} &
    \includegraphics[height=1.5cm]{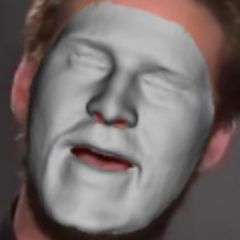} &
    \includegraphics[height=1.5cm]{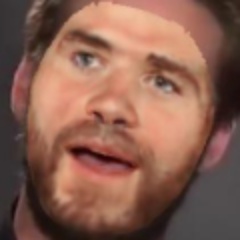} &
    \includegraphics[height=1.5cm]{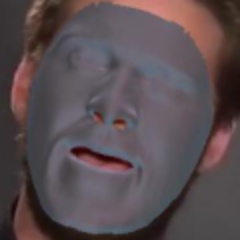} &
    \includegraphics[height=1.5cm]{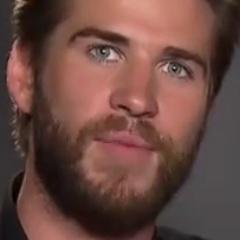} \\

    \end{tabular}
    }

    \caption{Qualitative comparison with Chaudhuri \etal \cite{chaudhuri2020personalized}.
    }
    \label{fig:qual_Chaudhuri_supp}
\vspace{-1em}
\end{figure}

\begin{figure*}[!t]
	\centering
    \includegraphics[width=0.9\linewidth]{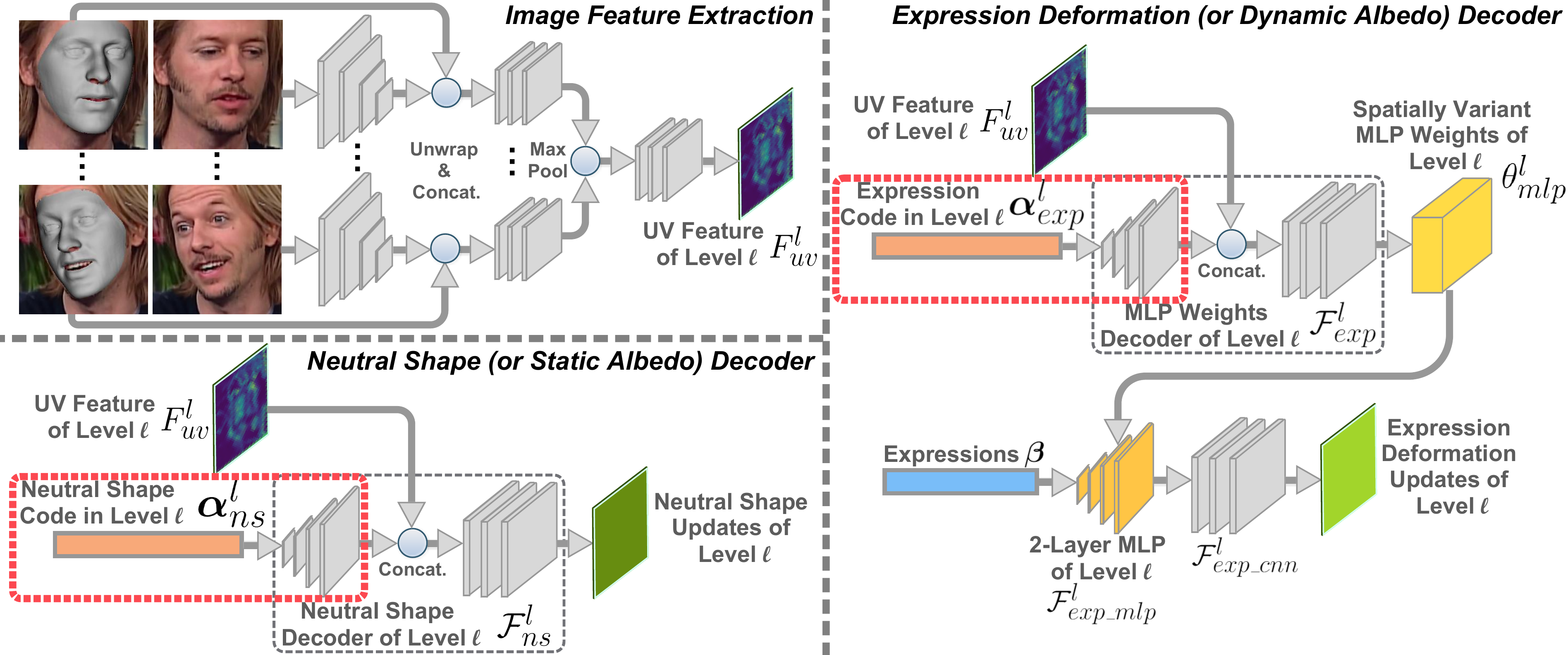}
    \caption{The illustration of modifications in the regression baselines (\secref{sec:regr}). Components inside the \textcolor[rgb]{1,0,0}{Red} boxes are removed.
    }
	\label{fig:rig_decoder_regr}
	\vspace{-1.0em}
\end{figure*}

\subsection{Detailed Loss Definitions (Main Paper Sec. 3.3)}
$L_{pose}$ is a pose-aware loss that supervises the per-image reconstruction, where we have two terms $L_{pose} = L_{dep\_v} + 0.025 L_{lan}$. In the depth-aligned vertex loss $L_{dep\_v}$, we first align the ground truth scan to the prediction in depth dimension since we do not estimate depth translation in our weak perspective camera model. Following \cite{bai2020deep}, we perform the depth alignment by adding the mean depth difference to the ground truth, then compute the $L_2$ distances of corresponding points between the prediction and the ground truth $L_{dep\_v} = \sum_i \sum_k \|\mathbf{v}_i^{gt,k} - \mathbf{\hat{v}}_i^k\|_2^2$ for all iterations and all levels. For the landmark loss $L_{lan}$, we adopt the same formulation as in \cite{bai2020deep}, which is a standard re-projection error. We use the $2$D locations of all landmarks from a $3$D detector (\ie first $2$ dimensions), and dynamic landmarks from a $2$D detector, as supervisions. We use different weights for different landmarks. For the landmarks of contour, eyebrow, and mouth, we use weight $10$, while for others (\ie eye, nose, and dynamic landmarks) we use weight $1$.

$L_{recon\_geo}$ is a geometry loss supervising the per-image reconstruction with two terms $L_{recon\_geo} = L_{den\_v} + 1000 L_{norm}$. We first rigidly align the prediction to the ground truth using dense correspondences. Then, the dense-aligned vertex loss $L_{den\_v}$ with the same form as $L_{dep\_v}$ and the normal loss $L_{norm} = \sum_i \sum_k (1 - \cos(\mathbf{n}_i^{gt,k}, \mathbf{\hat{n}}_i^k))$ are computed.

\subsection{Implementation Details}
Due to the incorrectness of the oversimplified image formulation and the memory limitation, we prevent the appearance relate energy and loss from influencing the shape estimation. More specifically, the photo-metric reconstruction energy $\mathbf{E}_{pho}$ only updates the albedo code $\bm{\alpha}_{alb}$, and the photo-metric loss $L_{recon\_pho}$ only trains the albedo related networks.

\section{Experiments}
\subsection{Training/Testing Data Consistency}
\label{sec:data_consist}
All testing data is the same for all methods (Bosphorus from \cite{bai2020deep}; BU3DFE from \cite{tewari2019fml}; NoW from \cite{sanyal2019learning}). \cite{tewari2019fml,chaudhuri2020personalized,sanyal2019learning} are self- or $2$D-supervise methods trained on in-the-wild images. \cite{tuan2017regressing} is a $3$DMM regression method thus trained with pre-fitted $3$DMM data. The differences on training data against us are due to the differences in the {\it methodology}. \cite{feng2018joint} is trained on pre-fitted $3$DMM data, while ours is trained with scans---a trade-off between data size and quality.

\subsection{Additional Qualitative Results}
\label{sec:qual}
We provide more qualitative results for per-image and video reconstructions as well as video retargeting.

\begin{figure}[h!]
    \centering
    \renewcommand{\arraystretch}{0.5}
    \resizebox{1.025\linewidth}{!}{
    \begin{tabular}{@{\hskip -1.5pt}c@{\hskip 1.5pt}c@{\hskip 1.5pt}c@{\hskip 1.5pt}c@{\hskip 1.5pt}c@{\hskip 1.5pt}c@{\hskip 1.5pt}c}
     & Input & Recon. & Albedo & Lighting & Shape & Bai20 \\
    \rotatebox[origin=l]{90}{\small Tewari19} &
    \includegraphics[height=1.5cm]{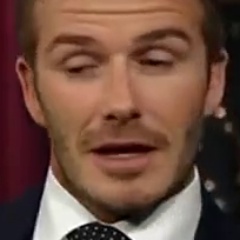} &
    \includegraphics[height=1.5cm]{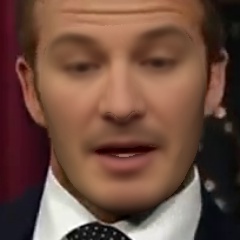} &
    \includegraphics[height=1.5cm]{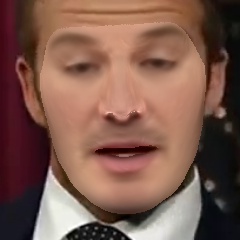} &
    \includegraphics[height=1.5cm]{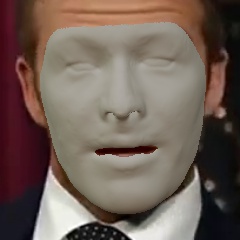} &
    \includegraphics[height=1.5cm]{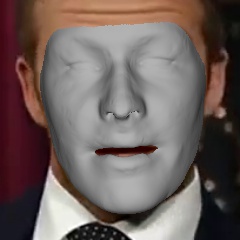} &
    \multirow{2}{*}[3ex]{\includegraphics[height=1.5cm]{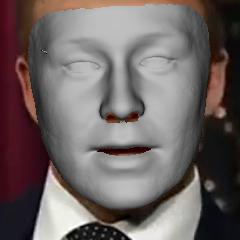}} \\
    \rotatebox[origin=l]{90}{\small Ours} &
    \includegraphics[height=1.5cm]{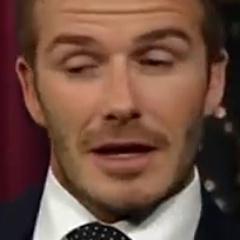} &
    \includegraphics[height=1.5cm]{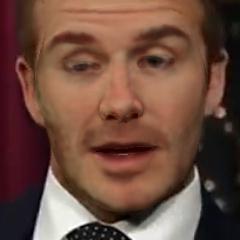} &
    \includegraphics[height=1.5cm]{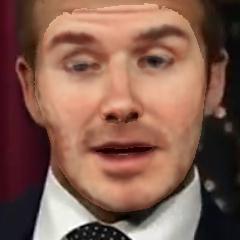} &
    \includegraphics[height=1.5cm]{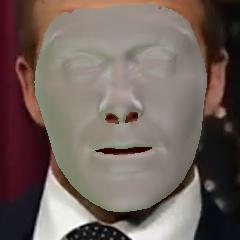} &
    \includegraphics[height=1.5cm]{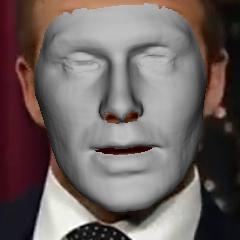} & \\
    
    \rotatebox[origin=l]{90}{\small Tewari19} &
    \includegraphics[height=1.5cm]{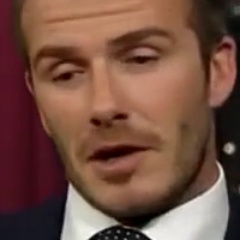} &
    \includegraphics[height=1.5cm]{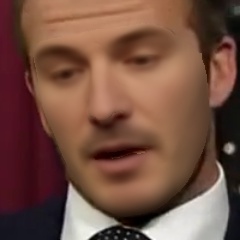} &
    \includegraphics[height=1.5cm]{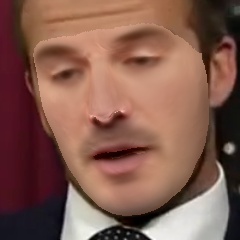} &
    \includegraphics[height=1.5cm]{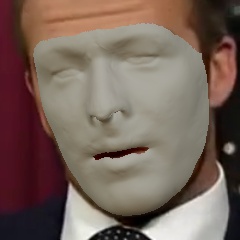} &
    \includegraphics[height=1.5cm]{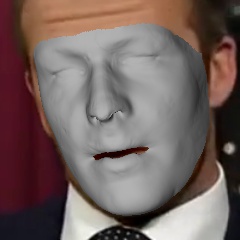} &
    \multirow{2}{*}[3ex]{\includegraphics[height=1.5cm]{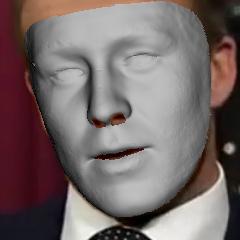}} \\
    \rotatebox[origin=l]{90}{\small Ours} &
    \includegraphics[height=1.5cm]{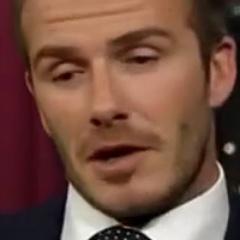} &
    \includegraphics[height=1.5cm]{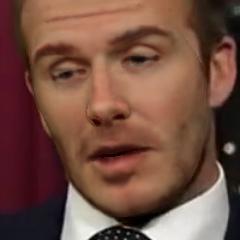} &
    \includegraphics[height=1.5cm]{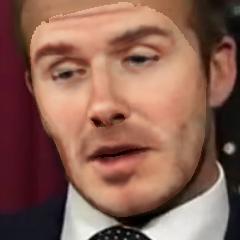} &
    \includegraphics[height=1.5cm]{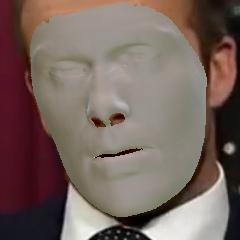} &
    \includegraphics[height=1.5cm]{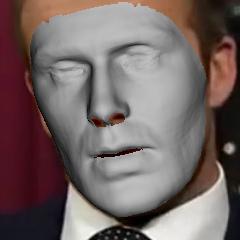} & \\
    
    \rotatebox[origin=l]{90}{\small Tewari19} &
    \includegraphics[height=1.5cm]{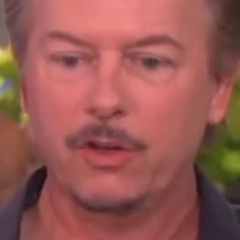} &
    \includegraphics[height=1.5cm]{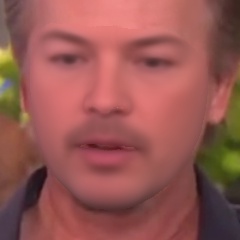} &
    \includegraphics[height=1.5cm]{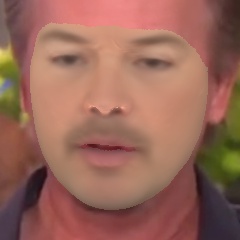} &
    \includegraphics[height=1.5cm]{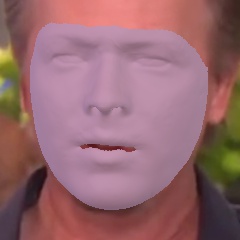} &
    \includegraphics[height=1.5cm]{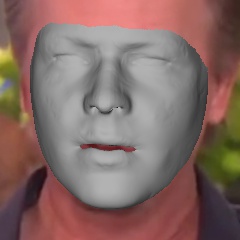} &
    \multirow{2}{*}[3ex]{\includegraphics[height=1.5cm]{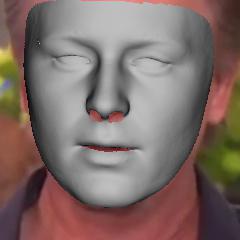}} \\
    \rotatebox[origin=l]{90}{\small Ours} &
    \includegraphics[height=1.5cm]{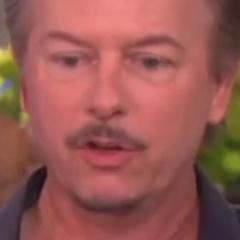} &
    \includegraphics[height=1.5cm]{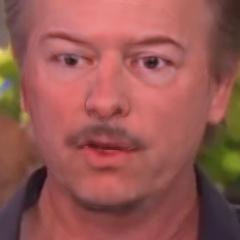} &
    \includegraphics[height=1.5cm]{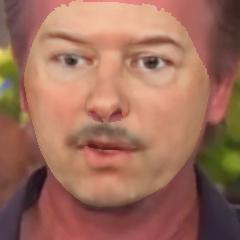} &
    \includegraphics[height=1.5cm]{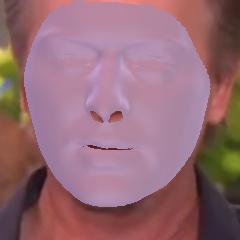} &
    \includegraphics[height=1.5cm]{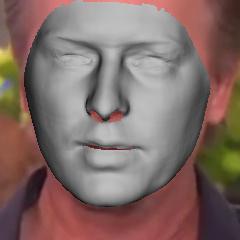} & \\
    
    \rotatebox[origin=l]{90}{\small Tewari19} &
    \includegraphics[height=1.5cm]{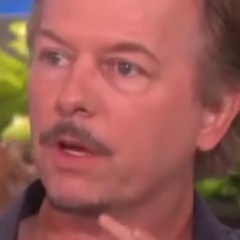} &
    \includegraphics[height=1.5cm]{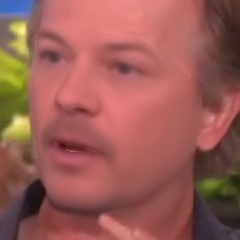} &
    \includegraphics[height=1.5cm]{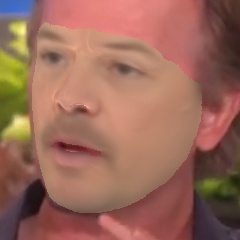} &
    \includegraphics[height=1.5cm]{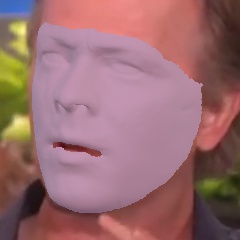} &
    \includegraphics[height=1.5cm]{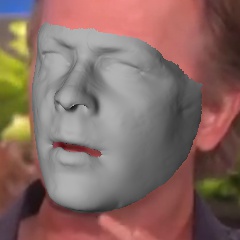} &
    \multirow{2}{*}[3ex]{\includegraphics[height=1.5cm]{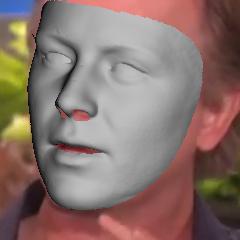}} \\
    \rotatebox[origin=l]{90}{\small Ours} &
    \includegraphics[height=1.5cm]{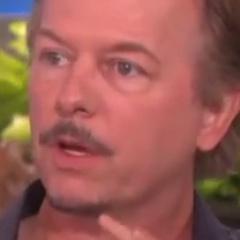} &
    \includegraphics[height=1.5cm]{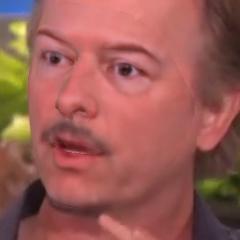} &
    \includegraphics[height=1.5cm]{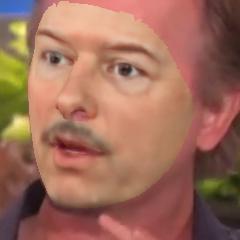} &
    \includegraphics[height=1.5cm]{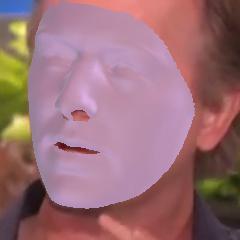} &
    \includegraphics[height=1.5cm]{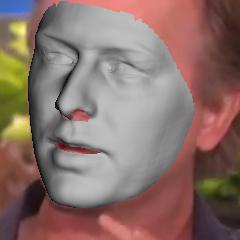} & \\
    
    \rotatebox[origin=l]{90}{\small Tewari19} &
    \includegraphics[height=1.5cm]{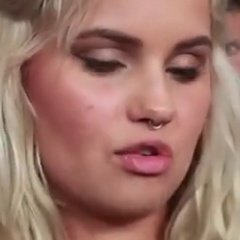} &
    \includegraphics[height=1.5cm]{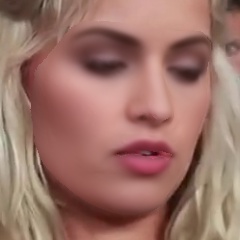} &
    \includegraphics[height=1.5cm]{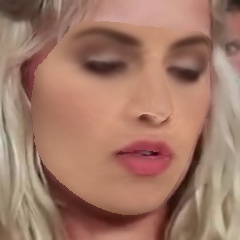} &
    \includegraphics[height=1.5cm]{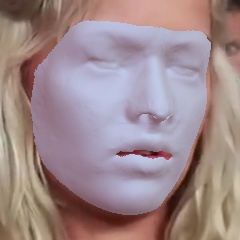} &
    \includegraphics[height=1.5cm]{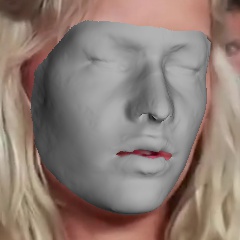} &
    \multirow{2}{*}[3ex]{\includegraphics[height=1.5cm]{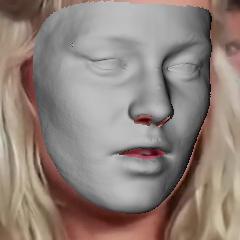}} \\
    \rotatebox[origin=l]{90}{\small Ours} &
    \includegraphics[height=1.5cm]{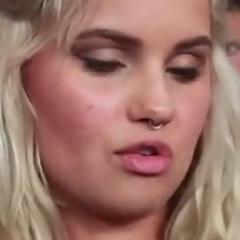} &
    \includegraphics[height=1.5cm]{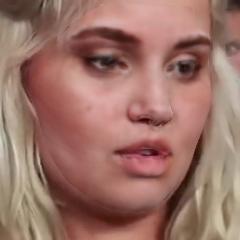} &
    \includegraphics[height=1.5cm]{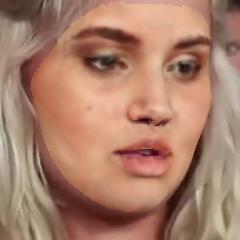} &
    \includegraphics[height=1.5cm]{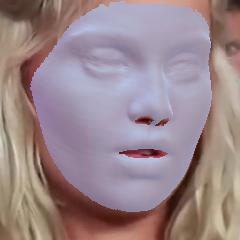} &
    \includegraphics[height=1.5cm]{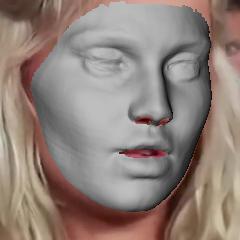} & \\
    
    \rotatebox[origin=l]{90}{\small Tewari19} &
    \includegraphics[height=1.5cm]{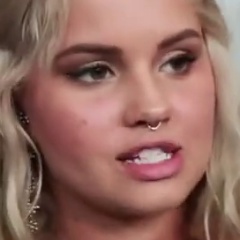} &
    \includegraphics[height=1.5cm]{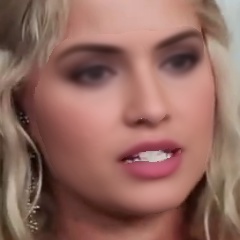} &
    \includegraphics[height=1.5cm]{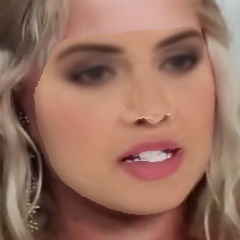} &
    \includegraphics[height=1.5cm]{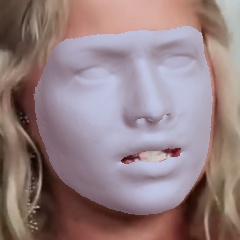} &
    \includegraphics[height=1.5cm]{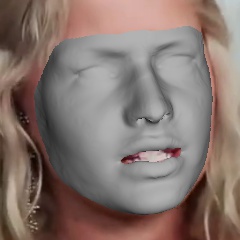} &
    \multirow{2}{*}[3ex]{\includegraphics[height=1.5cm]{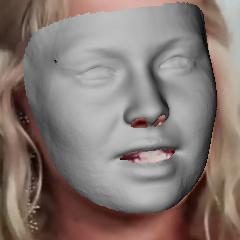}} \\
    \rotatebox[origin=l]{90}{\small Ours} &
    \includegraphics[height=1.5cm]{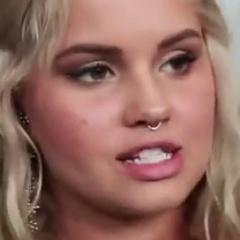} &
    \includegraphics[height=1.5cm]{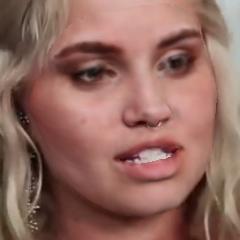} &
    \includegraphics[height=1.5cm]{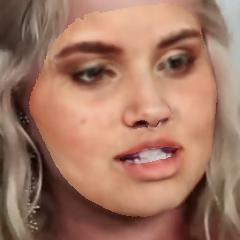} &
    \includegraphics[height=1.5cm]{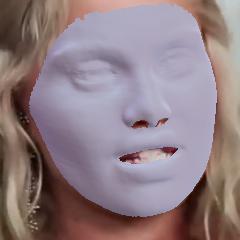} &
    \includegraphics[height=1.5cm]{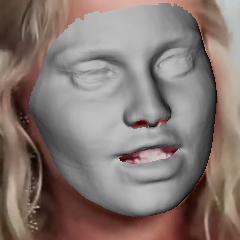} & \\
    
    \rotatebox[origin=l]{90}{\small Tewari19} &
    \includegraphics[height=1.5cm]{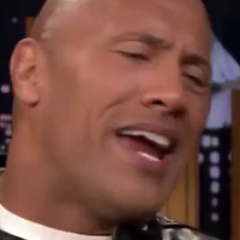} &
    \includegraphics[height=1.5cm]{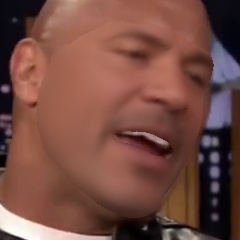} &
    \includegraphics[height=1.5cm]{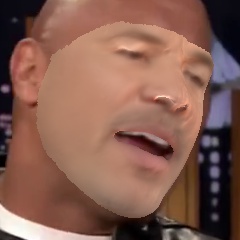} &
    \includegraphics[height=1.5cm]{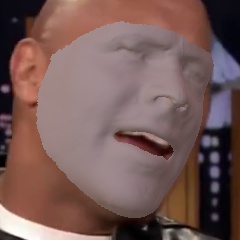} &
    \includegraphics[height=1.5cm]{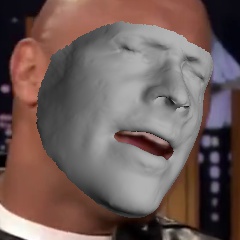} &
    \multirow{2}{*}[3ex]{\includegraphics[height=1.5cm]{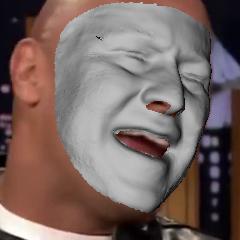}} \\
    \rotatebox[origin=l]{90}{\small Ours} &
    \includegraphics[height=1.5cm]{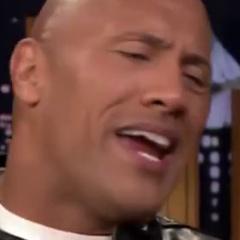} &
    \includegraphics[height=1.5cm]{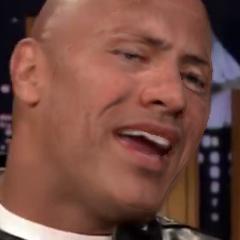} &
    \includegraphics[height=1.5cm]{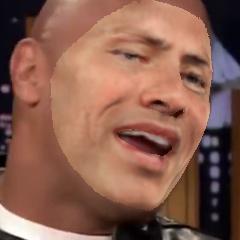} &
    \includegraphics[height=1.5cm]{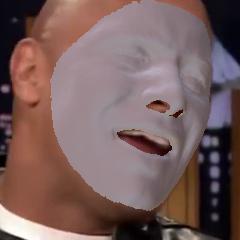} &
    \includegraphics[height=1.5cm]{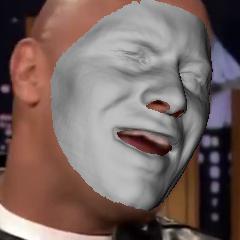} & \\
    
    \rotatebox[origin=l]{90}{\small Tewari19} &
    \includegraphics[height=1.5cm]{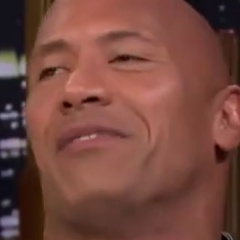} &
    \includegraphics[height=1.5cm]{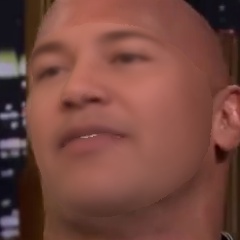} &
    \includegraphics[height=1.5cm]{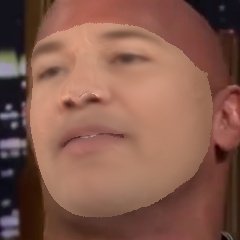} &
    \includegraphics[height=1.5cm]{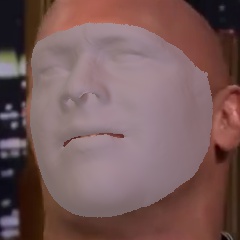} &
    \includegraphics[height=1.5cm]{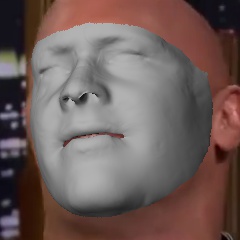} &
    \multirow{2}{*}[3ex]{\includegraphics[height=1.5cm]{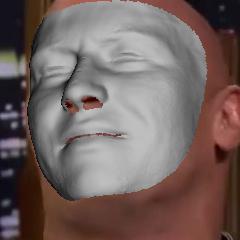}} \\
    \rotatebox[origin=l]{90}{\small Ours} &
    \includegraphics[height=1.5cm]{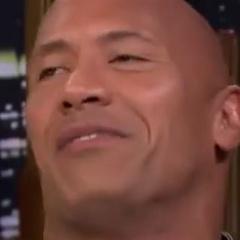} &
    \includegraphics[height=1.5cm]{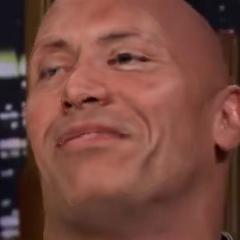} &
    \includegraphics[height=1.5cm]{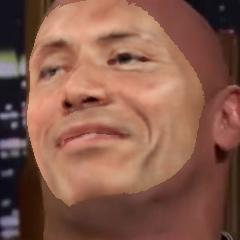} &
    \includegraphics[height=1.5cm]{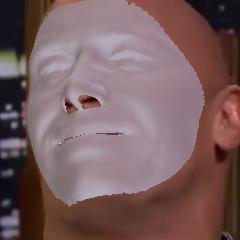} &
    \includegraphics[height=1.5cm]{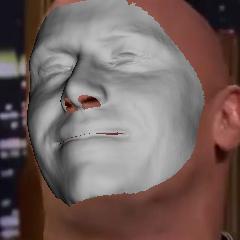} & \\

    \end{tabular}
    }

    \caption{Qualitative comparison with Tewari \etal \cite{tewari2019fml} and Bai \etal \cite{bai2020deep}.
    }
    \label{fig:qual_bai_fml_0}
\vspace{-1em}
\end{figure}

\begin{figure}[h!]
    \centering
    \renewcommand{\arraystretch}{0.5}
    \resizebox{1.025\linewidth}{!}{
    \begin{tabular}{@{\hskip -1.5pt}c@{\hskip 1.5pt}c@{\hskip 1.5pt}c@{\hskip 1.5pt}c@{\hskip 1.5pt}c@{\hskip 1.5pt}c@{\hskip 1.5pt}c}
     & Input & Recon. & Albedo & Lighting & Shape & Bai20 \\
    \rotatebox[origin=l]{90}{\small Tewari19} &
    \includegraphics[height=1.5cm]{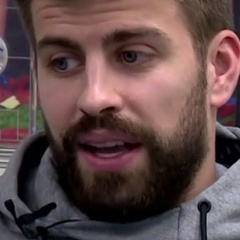} &
    \includegraphics[height=1.5cm]{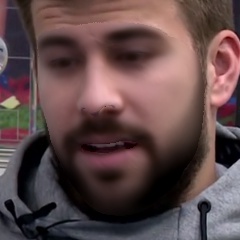} &
    \includegraphics[height=1.5cm]{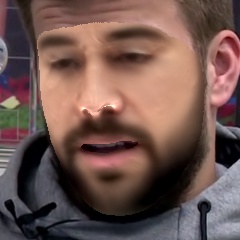} &
    \includegraphics[height=1.5cm]{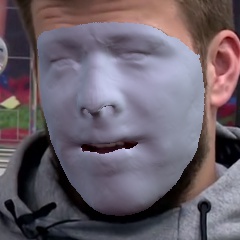} &
    \includegraphics[height=1.5cm]{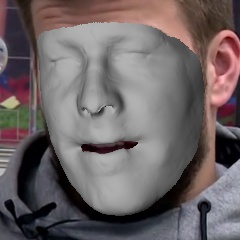} &
    \multirow{2}{*}[3ex]{\includegraphics[height=1.5cm]{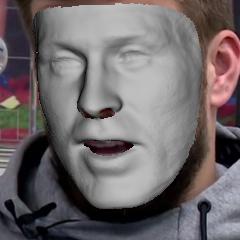}} \\
    \rotatebox[origin=l]{90}{\small Ours} &
    \includegraphics[height=1.5cm]{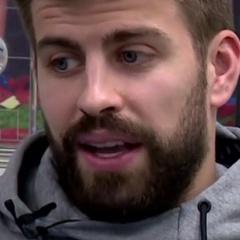} &
    \includegraphics[height=1.5cm]{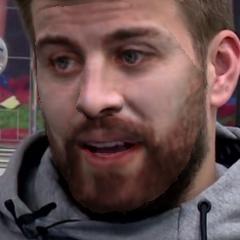} &
    \includegraphics[height=1.5cm]{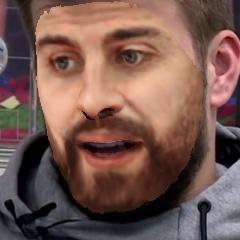} &
    \includegraphics[height=1.5cm]{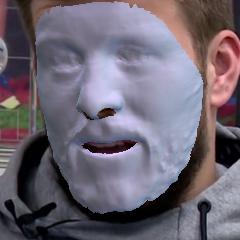} &
    \includegraphics[height=1.5cm]{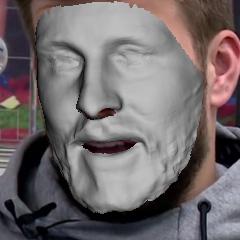} & \\
    
    \rotatebox[origin=l]{90}{\small Tewari19} &
    \includegraphics[height=1.5cm]{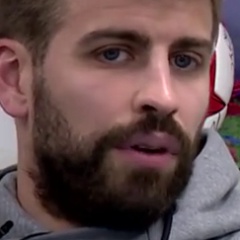} &
    \includegraphics[height=1.5cm]{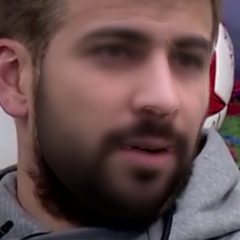} &
    \includegraphics[height=1.5cm]{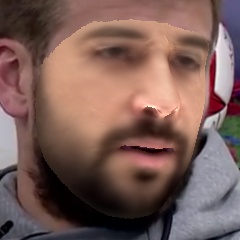} &
    \includegraphics[height=1.5cm]{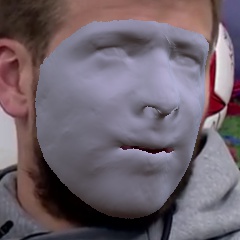} &
    \includegraphics[height=1.5cm]{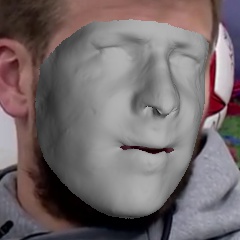} &
    \multirow{2}{*}[3ex]{\includegraphics[height=1.5cm]{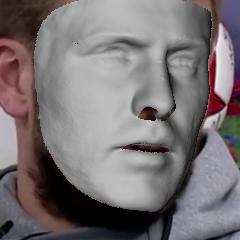}} \\
    \rotatebox[origin=l]{90}{\small Ours} &
    \includegraphics[height=1.5cm]{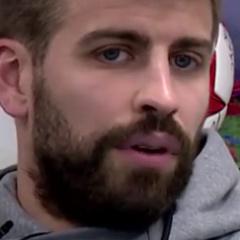} &
    \includegraphics[height=1.5cm]{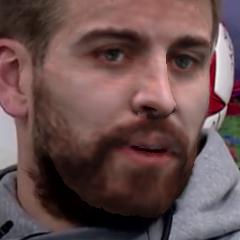} &
    \includegraphics[height=1.5cm]{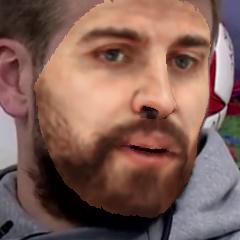} &
    \includegraphics[height=1.5cm]{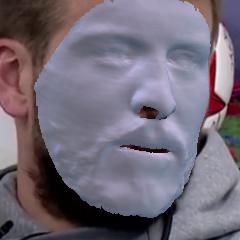} &
    \includegraphics[height=1.5cm]{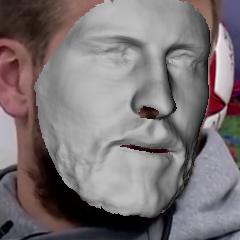} & \\
    
    \rotatebox[origin=l]{90}{\small Tewari19} &
    \includegraphics[height=1.5cm]{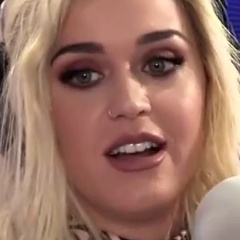} &
    \includegraphics[height=1.5cm]{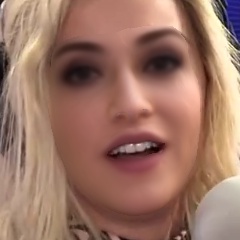} &
    \includegraphics[height=1.5cm]{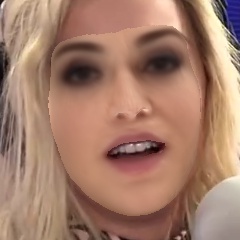} &
    \includegraphics[height=1.5cm]{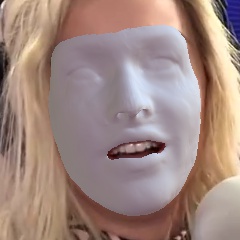} &
    \includegraphics[height=1.5cm]{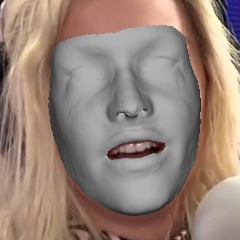} &
    \multirow{2}{*}[3ex]{\includegraphics[height=1.5cm]{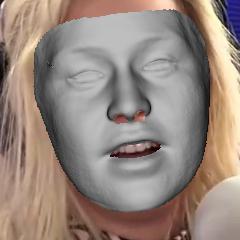}} \\
    \rotatebox[origin=l]{90}{\small Ours} &
    \includegraphics[height=1.5cm]{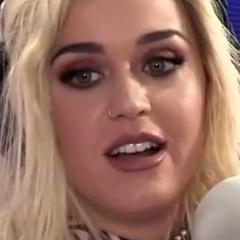} &
    \includegraphics[height=1.5cm]{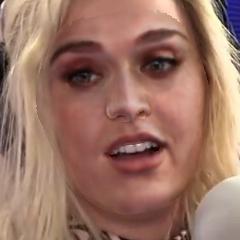} &
    \includegraphics[height=1.5cm]{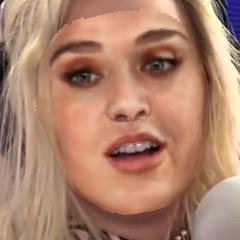} &
    \includegraphics[height=1.5cm]{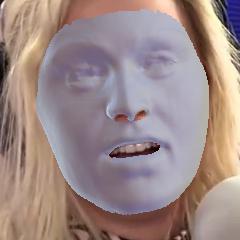} &
    \includegraphics[height=1.5cm]{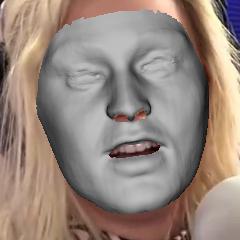} & \\
    
    \rotatebox[origin=l]{90}{\small Tewari19} &
    \includegraphics[height=1.5cm]{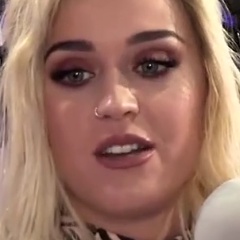} &
    \includegraphics[height=1.5cm]{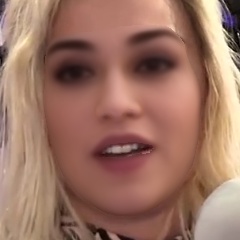} &
    \includegraphics[height=1.5cm]{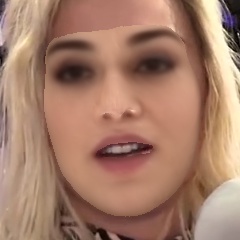} &
    \includegraphics[height=1.5cm]{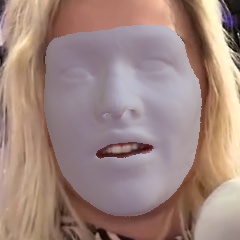} &
    \includegraphics[height=1.5cm]{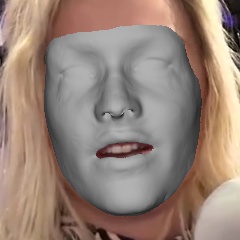} &
    \multirow{2}{*}[3ex]{\includegraphics[height=1.5cm]{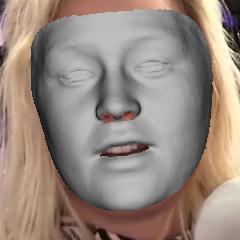}} \\
    \rotatebox[origin=l]{90}{\small Ours} &
    \includegraphics[height=1.5cm]{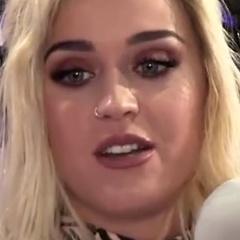} &
    \includegraphics[height=1.5cm]{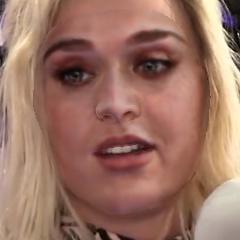} &
    \includegraphics[height=1.5cm]{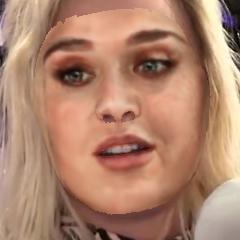} &
    \includegraphics[height=1.5cm]{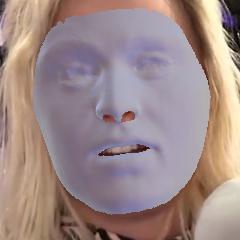} &
    \includegraphics[height=1.5cm]{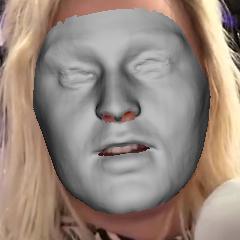} & \\
    
    \rotatebox[origin=l]{90}{\small Tewari19} &
    \includegraphics[height=1.5cm]{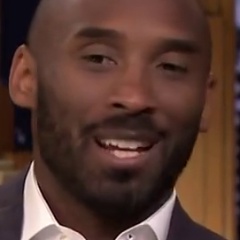} &
    \includegraphics[height=1.5cm]{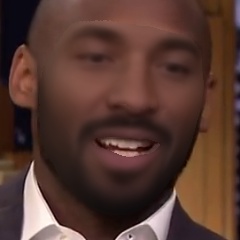} &
    \includegraphics[height=1.5cm]{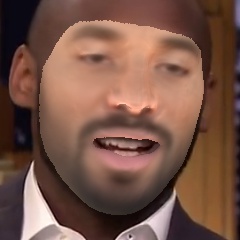} &
    \includegraphics[height=1.5cm]{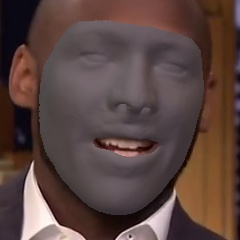} &
    \includegraphics[height=1.5cm]{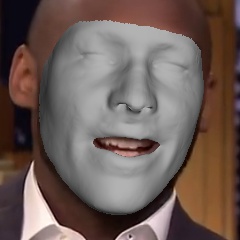} &
    \multirow{2}{*}[3ex]{\includegraphics[height=1.5cm]{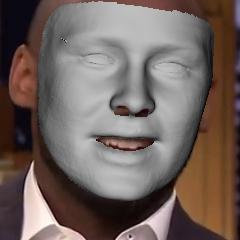}} \\
    \rotatebox[origin=l]{90}{\small Ours} &
    \includegraphics[height=1.5cm]{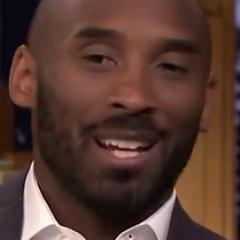} &
    \includegraphics[height=1.5cm]{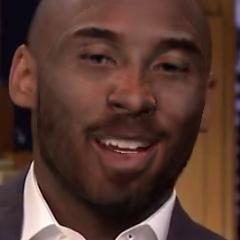} &
    \includegraphics[height=1.5cm]{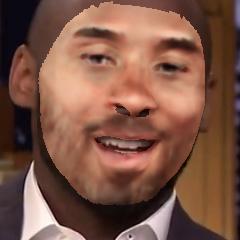} &
    \includegraphics[height=1.5cm]{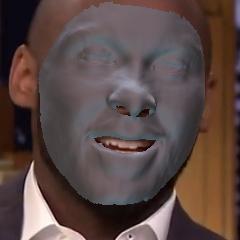} &
    \includegraphics[height=1.5cm]{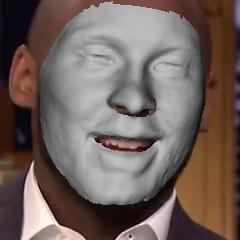} & \\
    
    \rotatebox[origin=l]{90}{\small Tewari19} &
    \includegraphics[height=1.5cm]{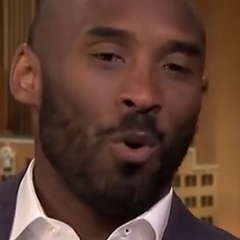} &
    \includegraphics[height=1.5cm]{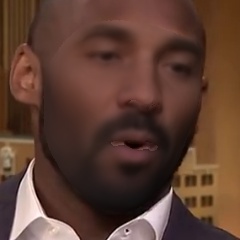} &
    \includegraphics[height=1.5cm]{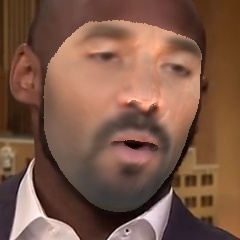} &
    \includegraphics[height=1.5cm]{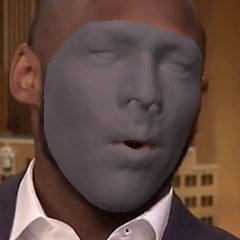} &
    \includegraphics[height=1.5cm]{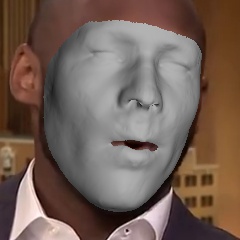} &
    \multirow{2}{*}[3ex]{\includegraphics[height=1.5cm]{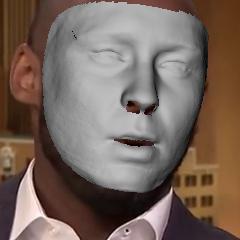}} \\
    \rotatebox[origin=l]{90}{\small Ours} &
    \includegraphics[height=1.5cm]{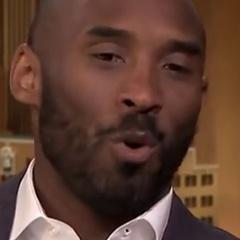} &
    \includegraphics[height=1.5cm]{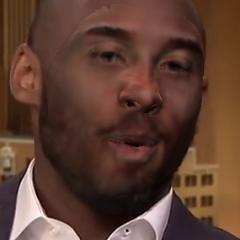} &
    \includegraphics[height=1.5cm]{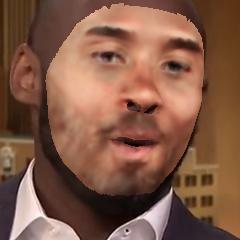} &
    \includegraphics[height=1.5cm]{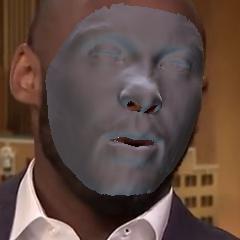} &
    \includegraphics[height=1.5cm]{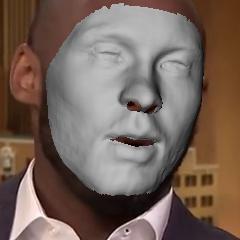} & \\
    
    \rotatebox[origin=l]{90}{\small Tewari19} &
    \includegraphics[height=1.5cm]{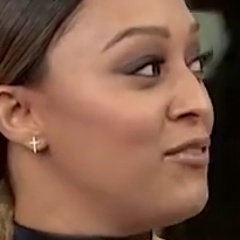} &
    \includegraphics[height=1.5cm]{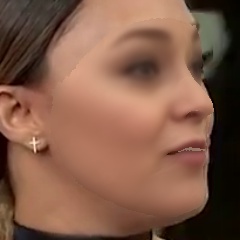} &
    \includegraphics[height=1.5cm]{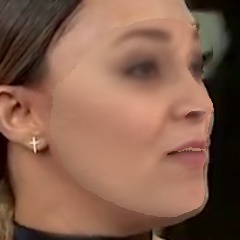} &
    \includegraphics[height=1.5cm]{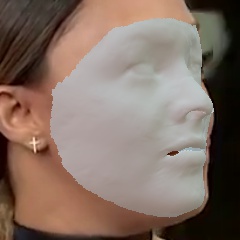} &
    \includegraphics[height=1.5cm]{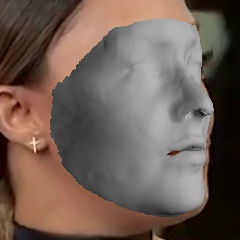} &
    \multirow{2}{*}[3ex]{\includegraphics[height=1.5cm]{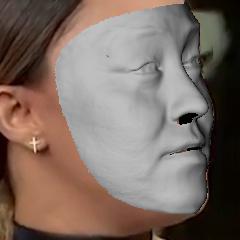}} \\
    \rotatebox[origin=l]{90}{\small Ours} &
    \includegraphics[height=1.5cm]{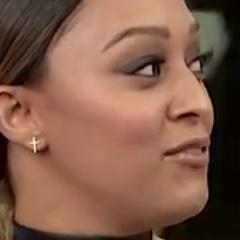} &
    \includegraphics[height=1.5cm]{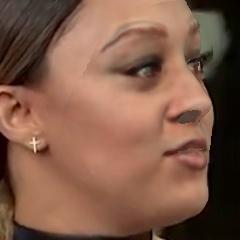} &
    \includegraphics[height=1.5cm]{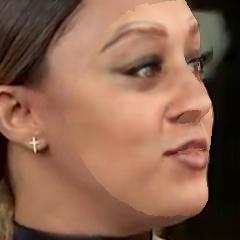} &
    \includegraphics[height=1.5cm]{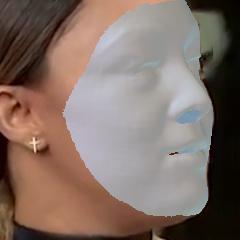} &
    \includegraphics[height=1.5cm]{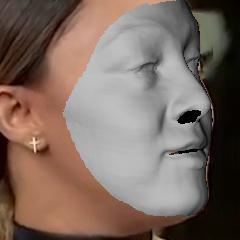} & \\
    
    \rotatebox[origin=l]{90}{\small Tewari19} &
    \includegraphics[height=1.5cm]{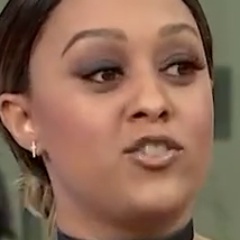} &
    \includegraphics[height=1.5cm]{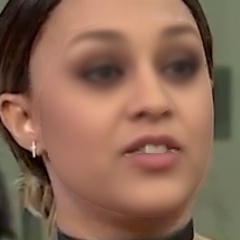} &
    \includegraphics[height=1.5cm]{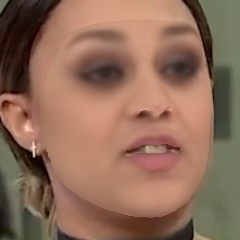} &
    \includegraphics[height=1.5cm]{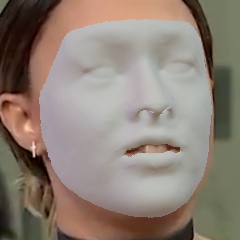} &
    \includegraphics[height=1.5cm]{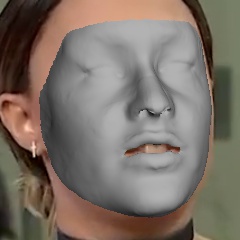} &
    \multirow{2}{*}[3ex]{\includegraphics[height=1.5cm]{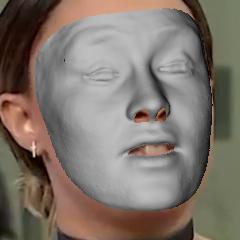}} \\
    \rotatebox[origin=l]{90}{\small Ours} &
    \includegraphics[height=1.5cm]{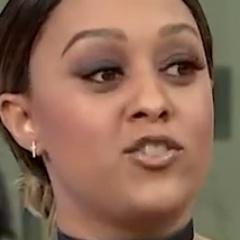} &
    \includegraphics[height=1.5cm]{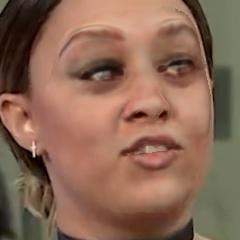} &
    \includegraphics[height=1.5cm]{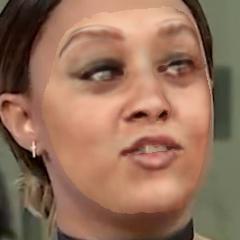} &
    \includegraphics[height=1.5cm]{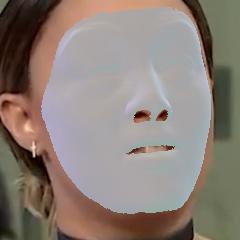} &
    \includegraphics[height=1.5cm]{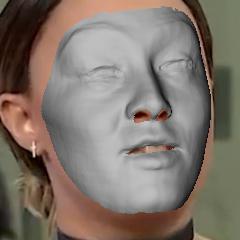} & \\

    \end{tabular}
    }

    \caption{Qualitative comparison with Tewari \etal \cite{tewari2019fml} and Bai \etal \cite{bai2020deep}.
    }
    \label{fig:qual_bai_fml_1}
\vspace{-1em}
\end{figure}

\subtitle{Per-image Reconstruction}
\figref{fig:qual_Chaudhuri_supp} shows the comparison with Chaudhuri \etal \cite{chaudhuri2020personalized}, where we get better geometries with more medium level details while having comparable textures. In \figref{fig:qual_bai_fml_0} and \figref{fig:qual_bai_fml_1}, we show comparisons with Tewari \etal \cite{tewari2019fml} and Bai \etal \cite{bai2020deep}. Our method produces more faithful shapes than Tewari \etal \cite{tewari2019fml} and Bai \etal \cite{bai2020deep} and higher resolution textures than Tewari \etal \cite{tewari2019fml}, though Tewari \etal \cite{tewari2019fml} achieves better albedo-illumination disentanglement.

\subtitle{Video Reconstruction and Retargeting}
For video reconstructions, we adopt the following strategy. Initially, we uniformly select $5$ frames from the video sequence and cache them. Given an incoming frame, we perform reconstruction using this frame together with the cached $5$ frames (\ie $6$ frames in total). Finally, the cached $5$ frames are updated to cover as large yaw angle range as possible. More specifically, we first sort the $6$ frames with estimated yaw angles. Then we discard the frame that has the smallest yaw angle difference with its neighbor (won't discard the first frame or the last one), and treat the rests as the updated $5$ cached frames. The estimated per-image parameters (\ie expressions, poses, and illuminations) are used for video retargeting. The supplementary video can be found at \href{https://youtu.be/vs7Kyv5rGas}{\textcolor[rgb]{0.9, .2, .2}{https://youtu.be/vs7Kyv5rGas}}.

Results on YouTube clips and videos from Bai \etal \cite{bai2020deep} and Chaudhuri \etal \cite{chaudhuri2020personalized} are included. On YouTube clips, our method achieves faithful reconstructions and reasonable retargeting results to various subjects. Compared with Bai \etal \cite{bai2020deep}, our method generates more stable reconstructions and additionally supports retargeting. Compared with Chaudhuri \etal \cite{chaudhuri2020personalized}, our method has superior shape quality that better reflects the personal characteristics, such as the round
chin instead of the sharp one from \cite{chaudhuri2020personalized} and the shape of the month, and achieves reasonable expression transfer results.

Originally, we planed to have a user study to quantitatively compare video retargeting results with Chaudhuri \etal \cite{chaudhuri2020personalized}, but we only have a demo video of \cite{chaudhuri2020personalized} that is not enough for a user study. As the code of \cite{chaudhuri2020personalized} is not publicly available, we contacted with the authors of \cite{chaudhuri2020personalized}. However, we were not able to get additional results at the end.

\subsection{3D Face Verification Details}
\label{sec:veri}
The $3$D face verification network is a ResNet34~\cite{he2016deep}, which takes in the UV representation of vertex positions and normals of the neutral shape (i.e. $6$ channels) and outputs an embedding. The network is trained with the contrastive loss~\cite{hadsell2006dimensionality} on the LFW training spilt under Restricted Configuration. We only train the first Conv \& BatchNorm layer and the last FC layer, while using weights pre-trained on ImageNet for all other layers. We also augment the input neutral shape with a random small rotation (\ie Euler angles sampled from $[-7.5\degree, 7.5\degree]$) during training for better robustness. 

\subsection{Regression Baseline Details}
\label{sec:regr}
To demonstrate the effectiveness of explicit optimization, we design regression baselines to compare with, where the components of the face rig are directly predicted by the neural network instead of being optimized. More specifically, we remove different parts of the rig code $\bm{\alpha}_{ns}$, $\bm{\alpha}_{exp}$ from the decoding process (\ie red boxes in \figref{fig:rig_decoder_regr}). Thus, the neutral shape or the network weights $\theta_{mlp}$ of the MLP are directly regressed without explicit optimization. For the neutral shape updates in level $1$, we additionally regress the $3$DMM coefficients from the UV feature map $F_{uv}$.

\subsection{Photometric Errors}
\label{sec:photo_err}
We test photometric errors on $220$ images selected by \cite{tewari2019fml} (on its website) from VoxCeleb2. 
As in \tabref{tab:photo_err}, although our method is trained on limited rendered images augmented with synthetic degradation, it performs on-par with the SOTA face modeling method \cite{tewari2019fml} that is trained on vast in-the-wild data. More discussions about the limitation in texture quality on in-the-wild data can be found in \secref{sec:limit}.

\begin{table}[!htb]
\caption{Photometric errors. Color value range: [0, 255).}
\label{tab:photo_err}
\setlength\tabcolsep{2.5pt}
\centering
  \begin{tabular}{c|c|c|c|c|c|c}
    & \multicolumn{3}{c|}{$L_{2,1}$ norm $\downarrow$} & \multicolumn{3}{c}{PSNR $\uparrow$} \\ \hline
    Methods & Ours & Ours(R) & [38] & Ours & Ours(R) & [38] \\
    \hline
    Mean & 14.81 & 14.12 & 14.05 & 29.66 & 30.06 & 30.10 \\
    STD & 4.58 & 4.43 & 3.46 & 2.53 & 2.65 & 2.22
  \end{tabular}
\end{table}

\subsection{Limitation}
\label{sec:limit}

\begin{figure}[!t]
	\centering
    \includegraphics[width=\linewidth]{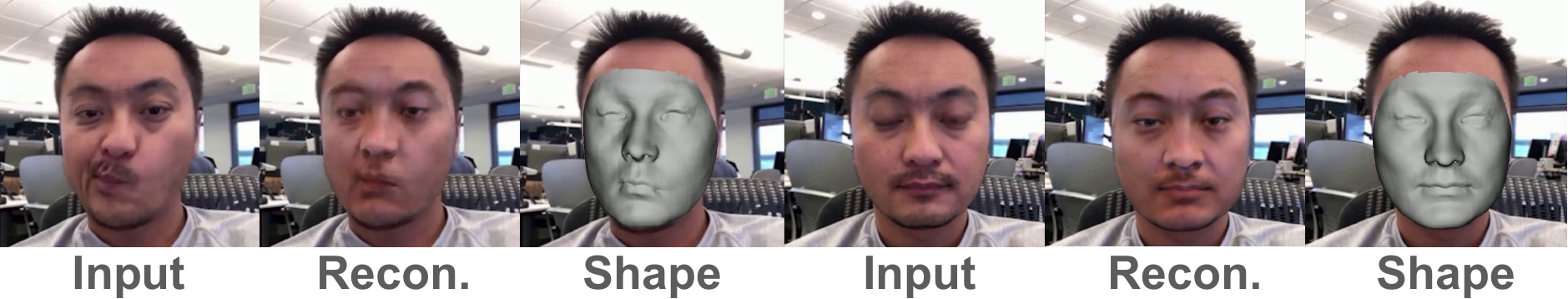}
    \caption{Failure cases of our method.}
	\label{fig:limitations}
	\vspace{-1.0em}
\end{figure}

Though our method achieves good reconstruction quality and reasonable retargeting results, we still observe some limitations on unusual expressions, eyelid motions, and the amplitude of transferred expressions. Currently our model cannot capture unusual expressions and eyelid motions well as in \figref{fig:limitations}, which could be due to the lack of training data since we use the Stirling/ESRC 3D face database \cite{esrc} for training where only $8$ expressions are included without unusual expressions and eyelid motions. Also, for some expressions (\eg the "frown" expression in 1:28 of the supplementary video), the amplitude of the transferred expression is slightly smaller than the source video, which could be due to the fact that the current space of the expression parameter $\bm{\beta}$ is automatically learned and not explicitly defined, such as blendshape coefficients. We leave this issue to future works.

Since our model is trained with rendered images (augmented with synthetic degradation for \textit{Ours(R)} version), its generalization ability is not perfect when applied on in-the-wild images, resulting in artifacts on textures (\eg making the face look dirty), which is also mentioned in Main Paper Sec.4.1. We believe our model could be improved with more realistically rendered training data and/or self-supervise learning directly on in-the-wild images.

\end{document}